\documentclass[10pt,twocolumn,letterpaper]{article}

 \usepackage[pagenumbers]{cvpr} 
\usepackage{tabu}
\usepackage{multirow}
\usepackage{subcaption}
\usepackage{adjustbox}
\usepackage[accsupp]{axessibility}
\usepackage{amsmath}

\usepackage{tikz}
\usepackage{pgfplots}
\pgfplotsset{compat=default}
\usetikzlibrary{pgfplots.groupplots}

\usepackage{adjustbox}
\usepackage{tabu}
\usepackage{multirow}
\usepackage{subcaption}

\newcommand{\TODO}[1]{\textbf{\color{red}[TODO: #1]}}
\renewcommand{\TODO}[1]{}

\definecolor{cvprblue}{rgb}{0.21,0.49,0.74}
\usepackage[pagebackref,breaklinks,colorlinks,citecolor=cvprblue]{hyperref}

\newcommand{\leon}[1]{\textcolor{black}{#1}}

\def\paperID{7687} 
\def\confName{CVPR}
\def\confYear{2024}

\title{Visual Prompting for Generalized Few-shot Segmentation: \\ A Multi-scale Approach}

\author{Mir Rayat Imtiaz Hossain$^{1,2}$ \qquad Mennatullah Siam$^{3,1}$  \qquad Leonid Sigal$^{1,2,4}$ \qquad James J. Little$^{1}$\\
$^1$University of British Columbia \qquad $^2$Vector Institute for AI \qquad $^3$ Ontario Tech University \\ \qquad $^4$Canada CIFAR AI Chair \\}

\begin{document}

\makeatletter
\DeclareRobustCommand\onedot{\futurelet\@let@token\@onedot}
\def\@onedot{\ifx\@let@token.\else.\null\fi\xspace}
\def\eg{\emph{e.g}\onedot} \def\Eg{\emph{E.g}\onedot}
\def\ie{\emph{i.e}\onedot} \def\Ie{\emph{I.e}\onedot}
\def\cf{\emph{cf}\onedot} \def\Cf{\emph{C.f}\onedot}
\def\etc{\emph{etc}\onedot} \def\vs{\emph{vs}\onedot}
\def\wrt{w.r.t\onedot} \def\dof{d.o.f\onedot}
\def\etal{\emph{et al}\onedot}

\def\latentregion{latent region}
\def\latentregionrep{latent token}

\makeatother
\maketitle



\begin{abstract}
The emergence of attention-based transformer models has led to their extensive use in various tasks, due to their superior generalization and transfer properties. Recent research has demonstrated that such models, when prompted appropriately, are excellent for few-shot inference. However, such techniques are under-explored for dense prediction tasks like semantic segmentation. In this work, we examine the effectiveness of prompting a transformer-decoder with learned visual prompts for the generalized few-shot segmentation (GFSS) task. Our goal is to achieve strong performance not only on novel categories with limited examples, but also to retain performance on base categories. We propose an approach to learn visual prompts with limited examples. These learned visual prompts are used to prompt a multiscale transformer decoder to facilitate accurate dense predictions. Additionally, we introduce a unidirectional causal attention mechanism between the novel prompts, learned with limited examples, and the base prompts, learned with abundant data. This mechanism enriches the novel prompts without deteriorating the base class performance. Overall, this form of prompting helps us achieve state-of-the-art performance for GFSS on two different benchmark datasets: COCO-$20^i$ and Pascal-$5^i$, without the need for test-time optimization (or transduction). Furthermore, test-time optimization leveraging unlabelled test data can be used to improve the prompts, which we refer to as transductive prompt tuning.

\end{abstract}

\vspace{-14pt}    
\section{Introduction}
\label{sec:intro}

\begin{figure}[t]
    \centering
    \includegraphics[width=0.41\textwidth]{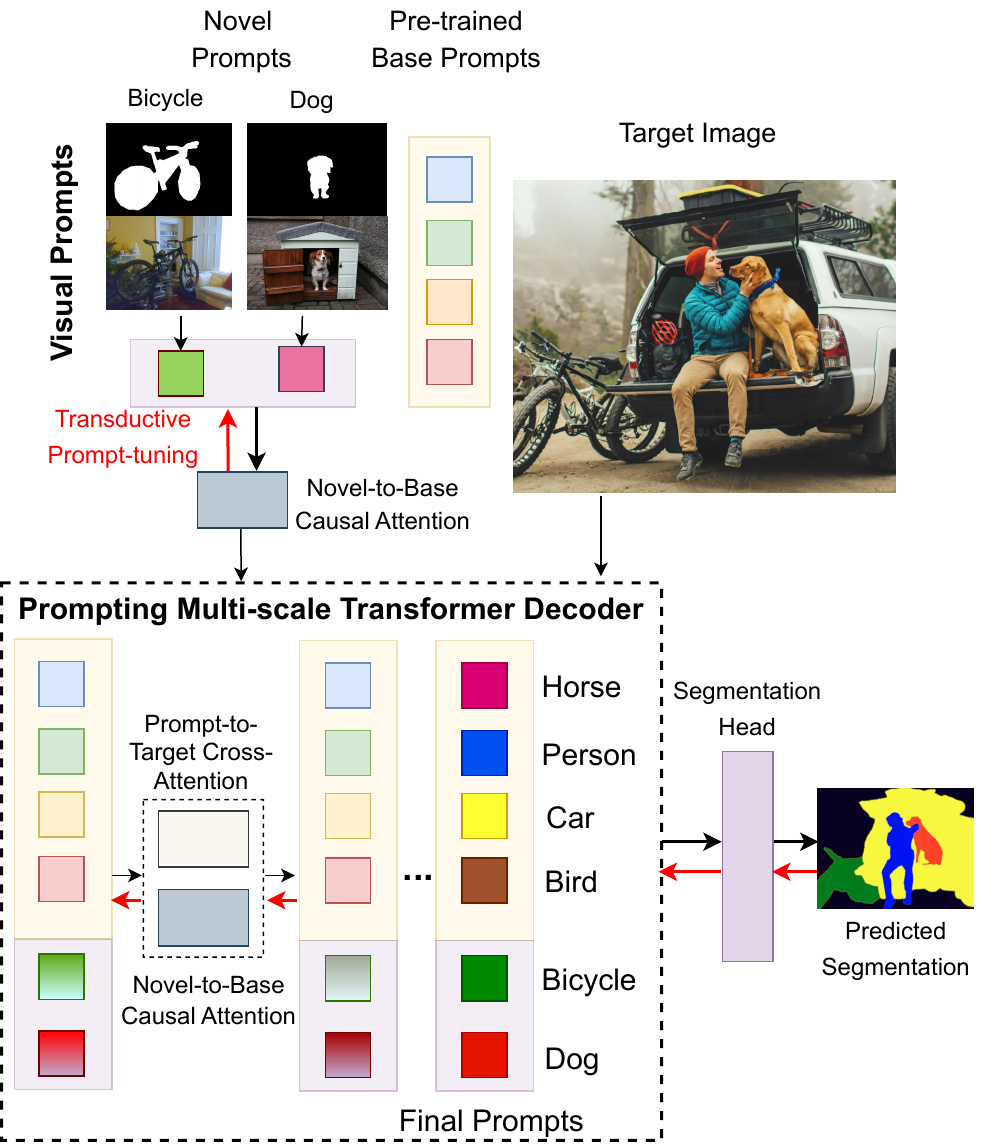}
    \vspace{-0.1in}
    \captionof{figure}{{\bf Overview of the Proposed Approach.} Prompting multi-scale transformer decoders for generalized few-shot segmentation. Our approach is a simple approach that allows test-time transductive prompt tuning (see red arrows).}
    \label{fig:intro}
    \vspace{-0.25in}
\end{figure}

The emergence of foundational models, trained on broad datasets, in natural language processing (e.g., GPT-3~\cite{brown2020language}) and in vision-language (e.g., CLIP \cite{radford2021learning}), exhibits powerful generalization and strong performance on multiple downstream tasks. These models have been adapted through different prompting techniques~\cite{wu2023cora,bulat2023fs} for them to be used in the few-shot scenario.

However, despite successes, including in localization \cite{wu2023cora, bulat2023fs, zhang2022feature}, prompting, which allows for few-shot demonstrations in dense prediction, and specifically in semantic segmentation tasks, is relatively under-explored.

Few-shot semantic segmentation \cite{liu2020part,zhang2019canet,wang2019panet,yang2020prototype,yang2020brinet,zhang2019pyramid} aims to segment {\em novel} (unseen) classes with few labelled training examples. Most state-of-the-art methods rely on meta-learning, to leverage abundant training data as a form of ``data augmentation" to construct many tasks that resemble test-time few-shot inference~\cite{yang2020prototype,min2021hypercorrelation}. In general, few-shot learning approaches can be classified as either {\em inductive} or {\em transductive}. 
Inductive approaches mainly rely on the training data, while transductive ones \cite{boudiaf2021few}, in addition, exploit unlabelled test data in an unsupervised manner to improve the performance
{\em e.g.}, by leveraging entropy priors over predictive class distributions.
However, both cases largely focus on the performance of the {\em novel} categories; which isn't particularly realistic. 
A recently proposed {\em generalized} few-shot segmentation (GFSS) setting defines a more realistic scenario where the goal is to perform well on {\em all} classes, novel and base~\cite{tian2022generalized}. This is considerably more challenging. 

Prompting has been proven to be effective for learning from few demonstrations, as seen in LLMs~\cite{brown2020language}. \textcolor{black}{It was also explored in visual prompt tuning~\cite{jia2022visual} to efficiently fine-tune vision transformers for new tasks.} We posit that prompting within a transformer-based~\cite{vaswani2017attention} architecture would similarly offer an effective and flexible mechanism for GFSS. This involves learning prompts that can be utilized to perform cross-attention with input image to make predictions. While learning prompts for base classes, with abundant data, is relatively straightforward, it becomes considerably more challenging for novel classes with few examples. Specifically, one must ensure that novel prompts learned from few samples are sufficiently distinct from the base prompts to avoid novel-base class misclassification.


To address the aforementioned challenges, we develop a simple yet highly effective visual prompting of transformer decoders for dense prediction at multiple scale that relies on novel-to-base causal\footnote{The term `causal'~\cite{oord2016wavenet} indicates its unidirectional nature.} attention \textcolor{black}{without meta-training}. \textcolor{black}{We look upon queries in DETR-style architectures as a form of visual prompts and devise a mechanism to initialize and learn novel prompts.} Then the novel-to-base causal attention allows base prompts to affect novel prompt representations, but not the other way around. This, intuitively, allows novel prompts to be repulsed and/or attracted towards their base counterparts. This attention is shared across scales (layers of the transformer) which, as we show, leads to more robust learning and improved performance. The multi-scale refinement of prompts helps interacting and reasoning between image features at multiple scales which in turn helps in better dense prediction. Finally, we extended this architecture in a transductive setting, where both novel and base prompts can be fine-tuned at test-time with respect to unsupervised objective to further improve performance. The proposed architecture and methodology is shown in Fig.~\ref{fig:intro}.

In summary, our contributions are:
\begin{itemize}
\item Designing a multi-scale visual prompting transformer decoder architecture for GFSS, featuring learnable prompts that allow the creation of new prompts for novel classes, initialized
through masked average pooling of support images (with their masks). 
\item Within this architecture, proposing and learning a multi-scale (shared) novel-to-base cross-attention mechanism between the novel and base prompts. 
\item Proposing a novel transductive prompt tuning, which allows the visual prompts to be tuned 
on test (unlabeled) images, hence the term {\em transductive}. 
\end{itemize}

\section{Related work}
\textbf{Few-shot Learning.} Few-shot learning, a longstanding concept in computer vision, serves as a paradigm to evaluate the data efficiency and adaptability of learning algorithms. It focuses on making predictions for {\em novel} classes with a few labeled examples, often by leveraging information from disjoint base classes for which there is abundant data. Few-shot learning has been extensively studied for image classification \cite{FinnICML2017,RaviICLR2017,RenICLR2018,snell2017prototypical,vinyals2016matching}. Most approaches can be categorized as either {\em transfer} or {\em meta}-learning-based. Transfer learning methods \cite{dhillon2019baseline} typically pre-train the model on base classes with abundant data and then fine-tune this model for novel classes with few samples. Meta-learning \cite{FinnICML2017,RaviICLR2017} approaches, on the other hand, simulate the inference stage during meta-training by sampling tasks of support and target sets. Meta-learning approaches can be sub-categorized into {\em gradient-based} and {\em metric-learning} based. Gradient-based approaches \cite{FinnICML2017,RaviICLR2017} learn the commonalities among different tasks. Metric-based (or prototype-based) methods \cite{snell2017prototypical,vinyals2016matching} learn image embeddings and compare the distances between them. Despite its popularity and strong performance, meta-learning is found to be sensitive to the meta-training setup~\cite{cao2019theoretical} and worse performance with domain shift~\cite{chen2019closer} is observed. 

Generalized few-shot learning, introduced in \cite{HariharanICCV2017} as a more realistic variant, allows query images from either base or novel categories. Recently, the transductive setting has been proven to be effective for few-shot learning~\cite{boudiaf2020information, qi2021transductive, boudiaf2021few, hajimiri2023strong, zhu2023transductive}. Transductive approaches exploit information of unlabelled test images during inference to improve classification~\cite{liu2018learning,nichol2018first,dhillon2019baseline,qiao2019transductive}. 
These approaches build on classic semi-supervised learning approaches, such as graph-based label propagation \cite{LiuICLR2019} or entropy minimization \cite{dhillon2019baseline}.
Recent variants of few-shot learning problems have also been established in other more granular domains, {\em e.g.}, detection \cite{Kang2019ICCV,WangICCV2019,YanICCV2019,WangICML2020} and segmentation.

\vspace{0.02in}
\noindent
{\bf Few-shot segmentation.}
Less common than classification, few-shot segmentation is emerging as the dense counterpart to the aforementioned approaches. Few-shot segmentation approaches can be categorized to {\em prototype-based} and {\em dense comparison-based} methods. Prototype-based methods mostly relied on extracted prototypes from the support set to guide the segmentation of the target image, either as a single prototype~\cite{zhang2019canet,wang2019panet}, part-aware~\cite{liu2020part} or a prototype mixture~\cite{yang2020prototype}. Dense comparison methods rely on dense comparisons between the few template images and the target image~\cite{yang2020brinet,zhang2019pyramid} with extension to multiscale~\cite{min2021hypercorrelation}. A recent approach proposed generalized few-shot segmentation (GFSS) task where the unlabelled target images are used to improve the segmentation of novel classes while maintaining the performance on base classes through knowledge distillation~\cite{hajimiri2023strong}. \textcolor{black}{A concurrent work~\cite{liu2023learning} on GFSS learns a set of orthogonal prototypes for base and novel categories, yet their GFSS setup is easier as they sample in their inference balanced base class images along with the novel ones, unlike the setup we follow~\cite{hajimiri2023strong}. Recent work~\cite{zhang2022feature} proposed a prompting strategy within a meta-learning framework using few examples. This approach is designed for 1-way segmentation only and performs prompting at a single scale unlike our proposed approach, a non-meta learning scheme, which performs multi-scale prompting and is designed for $k$-way GFSS.}

\vspace{0.02in}
\noindent
{\bf Prompting.} An emerging direction in learning with limited labelled data is to utilize foundation models~\cite{bommasani2021opportunities} that have strong generalization capabilities. These models, pre-trained on large-scale data, demonstrate strong generalization in few-shot settings~\cite{NEURIPS2022_960a172b}. Prompting is often used to adapt these foundation models to few-shot tasks, where it formulates the inference problem similarly to the training. Examples include auto-regressive modelling~\cite{NEURIPS2022_960a172b} where inference involves generating the next tokens after receiving the correct prompt.  For few-shot learning, prompts include few-shot demonstrations followed by the query. Prompting can employ discrete prompts~\cite{NEURIPS2022_960a172b} or learning continuous prompts~\cite{li2021prefix} through back-propagation. In computer vision, both vision-language prompting~\cite{NEURIPS2022_960a172b} and visual prompting~\cite{jia2022visual} of transformer-based models have shown success in various tasks, particularly in few-shot scenarios. However, effectively prompting models designed for localization and segmentation remains an open question.



\vspace{0.02in}
\noindent
{\bf Concurrent Work.}
A recent method~\cite{bulat2023fs} proposed prompting of transformer decoders for few-shot object detection. Although conceptually related, their approach is designed for a different task (detection) compared to ours (semantic segmentation), trained within a meta-learning framework unlike ours. Further, \cite{bulat2023fs} lacks novel-to-base attention and multi-scale prompt refinement which are at the core of our architectural design. A recent foundation model for semantic segmentation, SAM~\cite{kirillov2023segment}, exhibits strong performance when prompted by point, bounding box, or text.
However, it is not specifically designed for few-shot inference, potentially requiring fine-tuning with the support set and other non-trivial changes. 
More importantly, the use of extensive data to train the model makes it unsuitable for few-shot tasks since it has likely encountered many target classes during training. In addition to aforementioned differences, including multi-scale prompting and novel-to-base attention, our method is the first to explore transductive prompt-tuning where the visual prompts can be further optimized with the unlabelled target images.
\vspace{-0.05in}

\section{Problem Formulation} 
\begin{figure*}[t!]
\centering
\includegraphics[width=0.84\textwidth]{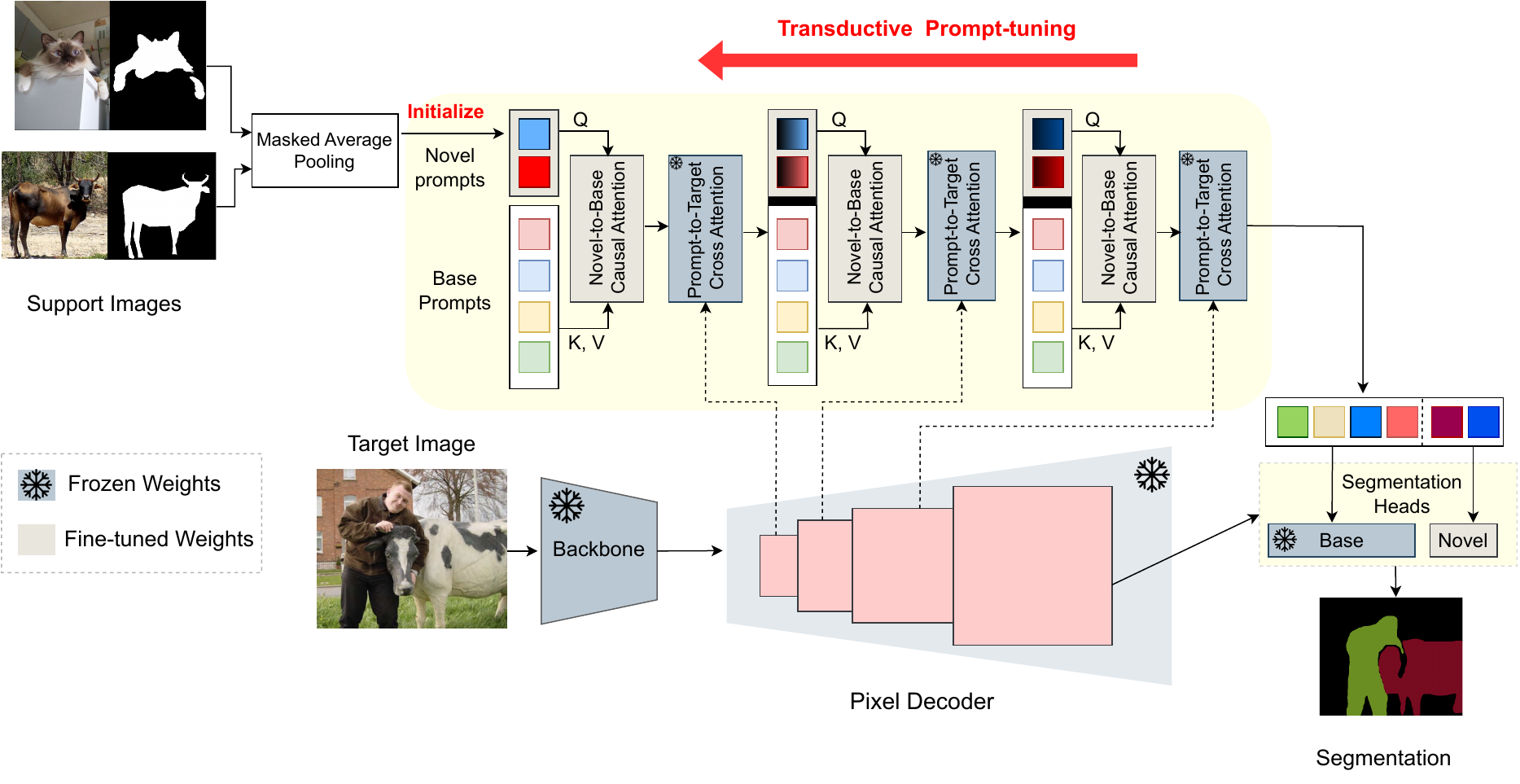} 
\vspace{-0.15in}
\caption{{\bf Detailed architecture of our proposed visual prompting of multiscale transformer decoder.} Our design initializes the novel visual prompts using the support set. This is followed by consecutive novel-to-base causal attention, $\mathcal{CA}$, and prompt-to-target features cross attention, $\mathcal{C}$, across scales. Note that causal attention uses shared weights across the scales and decoder layers. Our design allows for transductive fine-tuning of the visual prompts leveraging the unlabelled test image.}

\label{fig:model}
\vspace{-8pt}
\end{figure*}
\vspace{-2pt}
\noindent
\textbf{Classic Few-shot Segmentation.} In the classic few-shot segmentation scenario, the categories are divided into two sets of disjoint sets: \textbf{base} and \textbf{novel}. The problem is framed as an \textbf{$n$-way $k$-shot} segmentation task, where \textbf{$n$} is the number of novel classes and \textbf{$k$} is the number of examples available for each class. A typical solution involves {\em meta-learning} with two-stage training. In the meta-training stage, a support set is created with \textbf{$n$} randomly selected base classes, and \textbf{$k$} examples from each class are used to train the model. In the meta-testing stage, the model is evaluated on tasks with \textbf{$k$} examples from \textbf{$n$} novel classes. An alternative approach is non-meta learning, where a model is initially trained on abundant data for base classes, then frozen and fine-tuned for novel classes using few examples. The focus is on segmenting novel classes in test images without considering base class segmentation.

\noindent
\textbf{Generalized Few-shot Segmentation (GFSS).} {\em Generalized} few-shot segmentation, proposed by~\cite{tian2022generalized}, extends the classic few-shot setting. In this new setup the model must preserve the segmentation performance on base classes in addition to generalizing to novel classes. While during fine-tuning the support set (typically) only contains $k$-shot examples for each novel category, the model is still expected to make predictions of both novel and base classes on the test images. We follow the recent more practical setup of generalized few-shot segmentation~\cite{hajimiri2023strong} within a non-meta learning framework, details provided in Section~\ref{sec:implementation_details}.
\vspace{-0.10in}

\section{Method}

We perform prompting using a set of learnable visual prompts, where, intuitively\footnote{This may not be precisely true in practice, but is a reasonable abstraction for intuitive understanding.}, each prompt corresponds to a class \leon{embedding}. These learned visual prompts cross-attend with image features at multiple scales. Initially, we pre-train a base model with abundant examples of base classes, learning visual prompts for them. We then fix these base prompts and fine-tune new visual prompts for novel classes with a few examples. However, learning prompts from limited data is challenging because it leads to confusion between base and novel categories. To improve novel class generalization, we introduce a uni-directional causal attention mechanism between novel and base prompts. Intuitively, this allows better contextualization and repulsion / attraction  
between base and novel prompts. Section~\ref{sec:exps} demonstrates the empirical effectiveness of this mechanism.

Figure~\ref{fig:model} shows our proposed architecture. It centers around learnable embeddings called \emph{visual prompts}, which \leon{interact with} image features at multiple scales. Our design is inspired by Mask2former~\cite{cheng2022masked} with a couple of key differences: (i) the number of visual prompts aligns with the number of classes, each representing a class as a learned embedding; (ii) we adopt per-pixel classification, instead of mask classification, thereby eliminating the query classification component of \cite{cheng2022masked} and defining custom segmentation heads \leon{capable of making per-pixel predictions}.

\subsection{Visual Prompting Multiscale Transformers} ~\label{subsec:base}
\hspace{-4pt}\noindent {\bf Base Visual Prompts.} Given $B$ base classes, we define $B$ visual prompts each representing a class and initialize them randomly. These visual prompts first self-attend to learn the relationship between themselves and then are used to prompt the image features at multiple scales. The refinement of base prompts at each decoder layer is given by: 
\begin{equation}
    V_B^{(l)} = \mathcal{A}(V_B^{(l-1)}) + \mathcal{C}(V_B^{(l-1)}, F^{(l-1)}),
\end{equation}
\noindent
where $\mathcal{A}$ and $\mathcal{C}$ refer to multi-headed self and cross attention respectively; $V_B^{(l)} \in \mathbb{R} ^{B \times C} $ are the base visual prompts at layer $l$; $ F^{(l)} \in \mathbb{R}^{H_l W_l \times C_l}$ is the flattened image feature at layer $l$; $H_l$ and $W_l$ denote image dimensions at that layer; $C_l$ denotes the embedding dimension. 

After refining the prompts through multiple levels of transformer attention, a segmentation head, denoted as $\mathcal{H}(\cdot)$, is defined. This segmentation head takes the final refined prompts, $V_{B}^{(L)}$, and the highest resolution image features, $F^{(L)}$, as input, producing per-pixel per-class predictions  $O_{base} \in \mathbb{R}^{B\times H\times W}$ \leon{(see Section~\ref{sec:seghead})}.

\vspace{0.01in}
\noindent
{\bf Novel Visual Prompts.} 
After training on the base classes, we freeze all the layers of the model except the visual prompts representing the base classes. Given there are $N$ novel classes, we add $N$ new visual prompts. We further add a causal unidirectional attention module that allows the the novel prompts to be enhanced by the base prompts (see Section~\ref{subsec:base_to_novel}).  The output is passed to a segmentation head to make predictions for the novel classes.

The novel visual prompts are first refined by a unidirectional novel-to-base causal attention module. The enhanced novel prompts are then concatenated with the base prompts. Overall the concatenated prompts are refined by:
\begin{eqnarray} 
    V_N^{(l)} & = & \mathcal{CA}(V_B^{(l)}, V_N^{(l)}) \nonumber \\
    V_A^{(l)} & = & \left[ V_B^{(l)},  V_N^{(l)} \right] \nonumber \\
    V_A^{(l)}  & = & \mathcal{A}(V_A^{(l-1)} ) + \mathcal{C}(V_A^{(l-1)}, F^{(l-1)}), \nonumber
\end{eqnarray}
where $V_N^{(l)} \in \mathbb{R}^{N \times C} $ are novel prompts at layer $l$; $N$ is the total number of novel class and $C$ is the embedding size; $V_A^{(l)} \in \mathbb{R}^{(B+N)\times C} $ are all visual prompts including base and novel; $\left[.,.\right]$ is the concatenation operator. The operation $\mathcal{CA}$ denotes the causal novel-to-base attention and is explained in Section~\ref{subsec:base_to_novel}. 

Each of the $N$ novel visual prompts (representing a novel class) is initialized using masked average global pooling of the k-shot support images corresponding to the class masked by the ground truth binary masks. 
\begin{equation} \label{eq:init}
    V_n^{(0)} = \frac{1}{K}\sum\limits_{k=1}^K \frac{\sum\limits_{x, y} {M^k_n}(x, y) F^k_n(x, y)}{\sum\limits_{x, y} M^k_n(x, y)}, \forall n \in N
\end{equation}
where $V_n^{(0)} \in \mathbb{R}^{1 \times C}$ is the initial novel visual prompt for novel class $n$; $F^k_n$ denotes the image features of $k$-th shot for novel class;  $M_n^k$ denotes the corresponding ground-truth binary mask; $x, y$ are the spatial locations; $N$ is the total number of novel classes and $K$ is the total number of shots. 

\subsection{Novel-to-base Causal Attention}\label{subsec:base_to_novel}
A key component of our proposed approach is the uni-directional novel-to-base prompt causal\footnote{We use the term {\em causal} borrowing from its use in the look-ahead-attention-masking for decoder Transformers~\cite{vaswani2017attention} ({\em e.g.}, in language).} attention. The base visual prompts are learned with abundant examples; this is not the case for novel prompts. Learning novel prompts directly or having novel prompts impact the base ones could therefore lead to degradation of performance. However, having novel prompts aware of their base counterparts could be quite helpful. We hypothesize that causal attention between the novel visual prompts and base prompts could help contextualize novel class embeddings, reducing confusion with the base counterparts.  Motivated by this intuition, we propose a novel-to-base causal attention layer that is repeated at each scale (or layer of the transformer decoder); see Fig.~\ref{fig:model}.

The proposed causal attention module is a single layer cross-attention where the Key $K^{(l)}$, Value $V^{(l)}$ and $Q^{(l)}$ for layer $l$ of the decoder are given by: 
\begin{equation}
    Q^{(l)} = V_N^{(l)}  W^Q; ~~~~K^{(l)} = V_B^{(l)}  W^K; ~~~~V^{(l)} = V_B^{(l)} W^V \nonumber
\end{equation}
where $W^Q, W^K , W^V$ are weight matrices for $Q$, $K$ and $V$ respectively. The parameters $W^Q $, $W^K $, $W^V $ are shared at all the layers of this cross-attention module to reduce the number of trainable parameters during fine-tuning on novel classes to prevent overfitting and poor generalization. 

This uni-directional novel-to-base causal attention refines the novel visual prompts $V_N^{(l)}$ at each decoder layer $l$ by:
\begin{equation}
    V_N^{(l)} = \mathrm{softmax} (Q^{(l)} (K^{(l)})^\mathrm{T} ) V^{(l)} \nonumber
    \end{equation}
%
This results in a set of contextualized novel visual prompts.
\vspace{-0.05in}
\subsection{Segmentation Head}

\label{sec:seghead}
\noindent
{\bf Base Segmentation Head.} 
The base segmentation head, $\mathcal{H}_B$, during the base training in the proposed model is relatively simple and takes the form of a three layer MLP that projects a refined prompt into, effectively, a class prototype; computes the (dot product) similarity of each pixel feature with that prototype, and then applies softmax along the channel / class dimension to obtain per-pixel and per-class probability: 
\begin{eqnarray} \label{eq:base_output}
    O_{base} & = & \mathcal{H}_B(V_{B}^{(L)}, F^{(L)}) \nonumber \\ 
             & = &  \mathrm{MLP}\left(V_{B}^{(L)}\right) \cdot \left[F^{(L)}\right]^\mathrm{T},
\label{eq:baseseghead}
\end{eqnarray}
where $O_{base} \in \mathbb{R}^{B\times H\times W}$ and the class assignment can then be made by an $\mathrm{argmax}$ along the channel dimension. 


\vspace{0.01in}
\noindent
{\bf Novel Segmentation Head.}
While we could, in principle, use the same form of segmentation head during few-shot inference as that defined for the base classes (see Eq. \ref{eq:baseseghead}), simply fine-tuning it for novel classes, this proves ineffective in practice. 
Specifically, fine-tuning the MLP leads to significant overfitting given the few examples and lack of base class data during this step. 
So, instead, we resort to a simpler form of the novel segmentation head, $\mathcal{H}_N$, where we only learn a residual to the MLP learned during base training (which we keep fixed), mainly:
%
%
%
\begin{eqnarray}
    O_{novel} & = & \mathcal{H}_N(V_{N}^{(L)},  F^{(L)}) \nonumber \\
    & = & \left[ \mathrm{MLP}_{\bullet} (V_N^{(L)}) + W_N \right] \cdot \left[F^{(L)}\right]^\mathrm{T} , \nonumber
\end{eqnarray}
where $O_{novel}$ is output for novel classes and $\bullet$ designates that parameters of MLP are frozen and only $W_N \in \mathbb{R}^{N \times C}$ weight matrix is learned.
This weight matrix is initialized in the same manner as novel prompts, in Eq.~\ref{eq:init}. 

\vspace{-0.05in}
\subsection{Transductive Prompt Tuning} \label{subsec:transduction}

Recent approaches~\cite{boudiaf2021few, hajimiri2023strong} show the effectiveness of a good transductive setting for few-shot dense prediction tasks. Transductive approaches perform test-time optimization for each unlabeled test image and utilize specific characteristics and information of the test instances to improve prediction.

To demonstrate the flexibility of our approach, in this section we outline the process of adapting our model for transductive inference, improving GFSS performance through transductive fine-tuning of the visual prompts. We build on the transductive losses proposed in DIaM~\cite{hajimiri2023strong}  and we describe them here for completeness. 

The losses are designed to maximize the mutual information between the learned features and the corresponding predictions on the test image, which is achieved through maximizing: 
$\mathrm{H}(O) - \mathrm{H}(O | I)$, where $I, O$ are random variables associated with the pixels and predictions distributions respectively; $\mathrm{H}(O)$ is the marginal  and $\mathrm{H}(O | I)$ is the conditional entropy.
The conditional entropy $\mathrm{H}(O|I)$ involved is given by sum of cross-entropy loss over the supervised support set and entropy of the predicted probability given the test image $I$. The marginal entropy is given by KL divergence loss relying on an estimated region proportion prior following~\cite{boudiaf2021few}. Additional details are in \cite{hajimiri2023strong}.

Additionally, similar to~\cite{hajimiri2023strong}, we use knowledge-distillation loss to preserve base class performance, given by: 
\begin{equation}
    \mathcal{L}_{\mathrm{KD}} = \mathrm{KL} (O_{base}^{new} \ \ || \ \ O_{base}^{old}  ),
\end{equation}
where $O_{base}^{old}$ is the prediction probability of the base classes of the frozen model after base training; $O_{base}^{new}$  is the prediction probability of the base classes on the model with our additional components to handle novel categories that is being fine-tuned. These include the multi-scale causal novel-to-base cross attention and the novel prompts.

The overall transductive objective function is given by:
\begin{equation}  \label{eq:transduction}
    \mathcal{L}_{\mathrm{trans.}} = \alpha \mathrm{H}(O|I) - \mathrm{H}(O) + \gamma \mathcal{L}_{\mathrm{KD}},
\end{equation}
where $\alpha$ and $\gamma$ are balancing hyper-parameters.

We noticed that applying transduction from the first iteration underperforms, likely due to a poor initial estimate of the marginal distribution. To address this, we only apply per-pixel cross-entropy loss for a set number of iterations and then incorporate the remaining transductive losses from Eq.~\ref{eq:transduction} when we achieve a more accurate estimate of the marginal distribution.
\vspace{-6pt}

\section{Experiments}
\label{sec:exps}

\begin{table*}[t!]
\centering

 \begin{adjustbox}{max width=0.70\textwidth}
\begin{tabu}{@{}llcccccc@{}}
\tabucline[1pt]{-}
\multirow{1}{*}{} & \multirow{1}{*}{} & \multicolumn{6}{c}{} \\ 
 &  &  \multicolumn{6}{c}{\textbf{COCO-$20^i$}}\\

\multirow{2}{*}{Method} & \multirow{2}{*}{Learning} & \multicolumn{3}{c}{1-shot} & \multicolumn{3}{c}{5-shot}\\ 

\cmidrule(r{2pt}){3-5} \cmidrule(r{2pt}){6-8}

 & & Base  & Novel & Mean & Base & Novel & Mean  \\ \hline

CAPL~\cite{tian2022generalized} \ \tt{(CVPR22)}  & Inductive & 43.21 & 7.21 & 25.21 &  43.71 & 11.00 & 27.36\\

BAM~\cite{lang2022learning} \ \tt{(CVPR22)}  & Inductive & 49.84 & 14.16 & 32.00 &  49.85 & 16.63 & 33.24\\

DIaM (w/o trans.)$^*$ ~\cite{hajimiri2023strong}   \ \tt{(CVPR23)}  & Inductive & 42.69 & 15.32 & 29.00 &  38.47 & 20.87 & 29.67\\
POP$^\dagger$~\cite{liu2023learning} \ \tt{(CVPR23)}  & Inductive & 30.38 & 9.63 & 20.00 &  24.53 & 16.19 & 20.36\\
\textbf{Ours (w/o trans.)} & Inductive & \textbf{51.55} & \textbf{18.00} & \textbf{34.78} &  \textbf{51.59} & \textbf{30.06} & \textbf{40.83}\\

\hline
RePRI~\cite{boudiaf2021few} \ \tt{(CVPR21)}  & Transductive & 5.62 & 4.74 & 5.18 &  8.85 & 8.84 & 8.85\\

DIaM~\cite{hajimiri2023strong} \tt{(CVPR23)}  & Transductive & 48.28 & 17.22 & 32.75 &  48.37 & 28.73 & 38.55\\

POP$^\dagger$~\cite{liu2023learning} \ \tt{(CVPR23)}  & Transductive & \textbf{54.71} & 15.31 & 35.01 &  \textbf{54.90} & 29.97 & 42.44\\

\textbf{Ours (w/ trans.) }& Transductive & \underline{53.80} & \textbf{18.30} & \textbf{36.05} & \underline{53.81} & \textbf{31.14} & \textbf{42.48}\\

\tabucline[1pt]{-}
\multirow{1}{*}{} & \multirow{1}{*}{} & \multicolumn{6}{c}{} \\ 
\multirow{1}{*}{} & \multirow{1}{*}{} & \multicolumn{6}{c}{\textbf{PASCAL-$5^i$}} \\ 
\multirow{2}{*}{Method} & \multirow{2}{*}{Learning} & \multicolumn{3}{c}{1-shot} & \multicolumn{3}{c}{5-shot}  \\ 

\cmidrule(r{2pt}){3-5} \cmidrule(r{2pt}){6-8}

 & & Base  & Novel & Mean & Base & Novel & Mean  \\ \hline
CANeT~\cite{zhang2019canet} \ \tt{(CVPR19)}  & Inductive & 8.73 & 2.42 & 5.58 &  9.05 & 1.52 & 5.29\\
PFENET~\cite{tian2020prior} \ \tt{(TPAMI20)}  & Inductive & 8.32 & 2.67 & 5.50 &  8.83 & 1.89 & 5.36\\
PANET~\cite{wang2019panet} \ \tt{(ICCV19)}  & Inductive & 31.88 & 11.25 & 21.57 &  32.95 & 15.25 & 24.1\\

SCL~\cite{zhang2021self} \ \tt{(CVPR21)}  & Inductive & 8.88 & 2.44 & 5.66 &  9.11 & 1.83 & 5.47\\

MiB~\cite{cermelli2020modeling} \ \tt{(CVPR20)}  & Inductive & 63.80 & 8.86 & 36.33 &  68.60 & 28.93 & 48.77\\

CAPL~\cite{tian2022generalized} \ \tt{(CVPR22)}  & Inductive & 64.80 & 17.46 & 41.13 &  65.43 & 24.43 & 44.93\\

BAM~\cite{lang2022learning} \ \tt{(CVPR22)}  & Inductive & 71.60 & 27.49 & 49.55 &  71.60 & 28.96 & 50.28\\

DIaM (w/o trans.)$^*$ ~\cite{hajimiri2023strong}  \tt{(CVPR23)} & Inductive  & 66.79 & 27.36 & 47.08 & 64.05 & 34.56 & 49.31\\

POP$^\dagger$~\cite{liu2023learning} \ \tt{(CVPR23)}  & Inductive & 46.68 & 19.96 & 33.32 &  41.50 & 36.26 & 38.80\\

\textbf{Ours (w/o trans.)} & Inductive & \textbf{74.58} & \textbf{34.99} & \textbf{54.79} &  \textbf{74.86} & \textbf{50.34} & \textbf{62.60}\\

\hline
RePRI~\cite{boudiaf2021few} \ \tt{(CVPR21)}  & Transductive & 20.76 & 10.50 & 15.63 &  34.06 & 20.98 & 27.52\\

DIaM~\cite{hajimiri2023strong} \tt{(CVPR23)} & Transductive & 70.89 & 35.11 & 53.00 &  70.85  & 55.31 & 63.08\\

POP$^\dagger$~\cite{liu2023learning} \ \tt{(CVPR23)}  & Transductive & 73.92 & 35.51 & 54.72 &  74.78 & 55.87 & 65.33\\

\textbf{Ours (w/ trans.) }& Transductive & \textbf{76.39} & \textbf{39.83} & \textbf{58.11} &  \textbf{76.42} & \textbf{56.12} & \textbf{66.27}\\

\tabucline[1pt]{-}

\end{tabu}
\end{adjustbox}
\captionsetup{labelfont=bf}
\captionof{table}{{\bf Comparison to SoTA methods on COCO-$20^i$ and PASCAL-$5^i$.} The mIoU results are reported as the average across 5 different runs. All methods use ResNet-50 backbone. $^\dagger$ POP~\cite{liu2023learning} \textcolor{black}{uses an easier and less practical GFSS setup, e.g., balanced base class training in addition to support set during novel fine-tuning and uses a form of transduction; inductive setting is implemented by us following our setup}. Results of the rest of the methods are obtained from ~\cite{hajimiri2023strong} except the inductive setting of DIaM$^*$, which we implemented ourselves. The \textbf{w/o trans.},\textbf{ w/ trans.} denote without and with transduction respectively. }
\vspace{-0.1in}
 \label{tab:coco}
\end{table*}

\begin{table}[t]
     \centering
     \aboverulesep=0ex
     \belowrulesep=0ex
    \begin{adjustbox}{max width=\columnwidth}
    \begin{tabular}{lc|c|c|c|c|c}
        \toprule
        
           &  \multirow{2}{*}{Causal Attention} & \multirow{2}{*}{Prompt Initialization} & \multirow{2}{*}{Transduction} & \multicolumn{3}{c}{mIoU}\\\cmidrule(r{2pt}){5-7}
          &  & & &  Base & Novel & Mean \\
        \midrule
      \textbf{(1)} & None & Random & No & 50.18 & 11.22 & 30.70\\ 

    \textbf{(2)} & None  &   Masked Pooling    &   No  & 50.43 & 11.31 &  30.87  \\

   \textbf{(3)}  &  First Layer Only  &   Masked Pooling    &  No  & 51.53 & 12.26 &  31.90  \\
   \textbf{(4)} &   Separate Weights  &  Masked Pooling   &    No  & 51.06 & \underline{18.05} & 34.56   \\ 

    \textbf{(5)} & Shared Weights  &  Random   &  No  & 50.75 & 17.41 &  34.08   \\
    \textbf{(6)} & Shared Weights  &  Masked Pooling   &    No  & \underline{51.55} & 18.00 &  \underline{34.78}   \\ 
      
   \textbf{(7)} &  Shared Weights  &  Masked Pooling  &  Yes & \textbf{53.80} & \textbf{18.30} &  \textbf{36.05}   \\  
        \bottomrule
      \end{tabular}
  \end{adjustbox}
  \vspace{-4pt}
     \caption{{\bf Ablation Study on COCO-$20^i$.} The 1-shot performane, averaged over all folds.}
     \label{tab:ablation}
\vspace{-1.5em}
\end{table}

\subsection{Dataset and Implementation Details}
\label{sec:implementation_details}
\noindent
{\bf Datasets.} We evaluate our method on two few-shot segmentation benchmarks~\cite{tian2022generalized}: COCO-$20^i$ and PASCAL-$5^i$. Following~\cite{hajimiri2023strong} we report average performance of our model on 10K test images for COCO-$20^i$ and on all test images on PASCAL-$5^i$. The results are reported as an average across five different runs on four different splits. 

\vspace{0.01in}
\noindent
{\bf Data Pre-processing.} We adopt the same setup as DIAM~\cite{hajimiri2023strong}. During base training, images containing both base and novel categories are kept, but the pixels corresponding to novel classes are relabelled as background. This approach, as noted by~\cite{hajimiri2023strong}, presents challenges due to potential ambiguity when the base model predicts novel categories as background. In fine-tuning, only novel classes are labeled in the support set, with base classes marked as background. To avoid unfair shot increases, images with multiple novel categories are excluded from the support set.

\vspace{0.01in}
\noindent
{\bf Evaluation Protocol.} We evaluate performance using the standard mean intersection-over-union (mIoU) metric. Following~\cite{hajimiri2023strong}, the mean score for generalized few-shot segmentation is computed as the average between \emph{base} and \emph{novel} mIoUs. This approach addresses bias in datasets like PASCAL-$5^i$ and COCO-$20^i$, where base classes outnumber novel classes threefold.

\vspace{0.01in}
\noindent
{\bf Implementation Details.} We use PSPNet~\cite{zhao2017pyramid} with Resnet-50~\cite{he2016deep} as feature extractor. We utilize MSDeformAttn-based pixel decoder~\cite{zhu2020deformable} from Mask2Former~\cite{cheng2022masked}. The dimension $C$ for visual prompts is chosen to be $256$. Contrary to \cite{cheng2022masked}, the base model is trained with 
per-pixel cross-entropy loss. We use AdamW optimizer~\cite{loshchilov2018decoupled} with base learning rate of $1 \times 10^{-4}$, a weight decay of $0.05$ and a batch size of $16$. 
Following~\cite{lang2022learning, hajimiri2023strong}, we train base model for 20 epochs for COCO-$20^i$ and 100 epochs for PASCAL-$5^i$. 

During novel fine-tuning, we freeze the parameters of the feature extractor, pixel decoder, and base segmentation head, and only optimize the parameters of the visual prompts, novel segmentation head, and novel-to-base causal attention module. For the inductive case, the model is fine-tuned with per-pixel cross-entropy loss for 100 iterations. For the transductive setting, the model is fine-tuned for 40 iterations with only cross-entropy loss and then fine-tuned for 60 more iterations with full transductive losses mentioned in Eq.~\ref{eq:transduction}. The hyper-parameters  $\alpha$ and $\gamma$ are set to be 
100 and 25 respectively. We use AdamW optimizer with a learning rate of $5 \times 10^{-3}$ and a weight decay of 0.05 for fine-tuning. 
\subsection{Results}
\vspace{-4pt}
\noindent
{\bf Comparison with the State-of-The-Art.} Table~\ref{tab:coco} compares our approach in both {\em inductive} and {\em transductive} setting against the state-of-the-art. As observed, our model outperforms existing approaches. Interestingly, our approach, in an inductive setting (where optimization is performed only using a supervised few-shot support set, and evaluation is conducted on all test images), outperforms transductive approaches like DIaM~\cite{hajimiri2023strong} for both 1-shot and 5-shot cases on the COCO-$20^i$ dataset and for the 1-shot case on Pascal-$5^i$. This is notable despite transductive approaches requiring test-time optimization for each unlabeled test image separately, leading to significantly longer inference times.

For a fair comparison in the {\em inductive} setting, we fine-tuned DIaM without the transduction losses. We also fine-tuned POP~\cite{liu2023learning} in our inductive setup. POP~\cite{liu2023learning} randomly samples balanced base class image, making it easier for them to retain base class information and uses a form of transduction where their model relabels novel class pixels initially labelled as background. We observe that the performance of POP~\cite{liu2023learning} degrades substantially, especially for the base classes in our inductive setting. 

Our inductive approach has 5.78\% higher mIoU for 1-shot and 11.16\% higher mIoU for 5-shot on COCO-$20^i$ than inductive DIaM. For PASCAL-$5^i$, the improvement from inductive DIaM is 7.71\% for 1-shot and 13.29\% for 5-shot. In fact, compared to the best inductive GFSS approach BAM~\cite{lang2022learning}, our inductive setting achieves an improvement of approximately 2.78\% and 7.59\% for 1-shot and 5-shot respectively on COCO-$20^i$ and 5.24\% and 12.32\% for 1-shot and 5-shot respectively on PASCAL-$5^i$. Our method, in the inductive setting, even outperforms transductive DIaM~\cite{hajimiri2023strong} by 2.03\% and 2.28\% for 1-shot and 5-shot respectively on COCO-$20^i$ and by 1.79\% for 1-shot on PASCAL-$5^i$

On applying {\em transductive} fine-tuning to our method, overall mIoU increases considerably, outperforming transductive DIaM~\cite{hajimiri2023strong} and POP~\cite{liu2023learning} on both 1-shot and 5-shot inference for both datasets. The improvement over DIaM for COCO-$20^i$ is 3.3\% for 1-shot and 3.93\% for 5-shot. For PASCAL-$5^i$, our transductive setting outperforms DIaM by 5.11\% and 3.15\% for 1 and 5-shots.

\begin{figure}[t]
    \centering
    \includegraphics[width=0.5\textwidth]{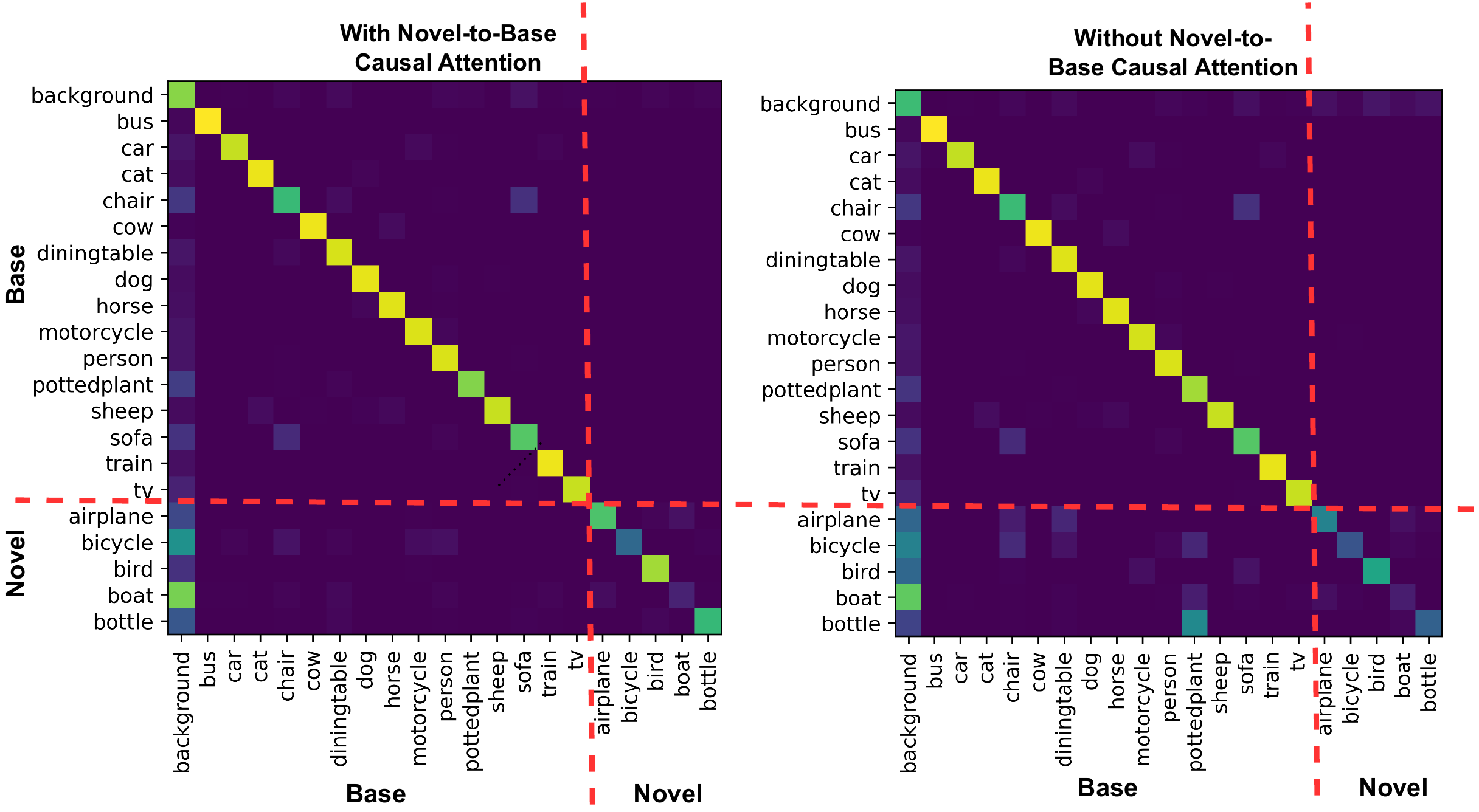}
    \vspace{-0.1in}
    \captionof{figure}{{\bf Confusion Matrix for:} (left) model with Novel-to-Base Causal Attention; (right) model without Novel-to-Base Causal Attention. \leon{Note that Novel-to-Base Causal Attention reduces confusion between novel and base categories (bottom-left block).}}
    \label{fig:confusion_matrix}
    \vspace{-0.25in}
\end{figure}

\begin{figure}[t]
    \centering
    \includegraphics[width=0.5\textwidth]{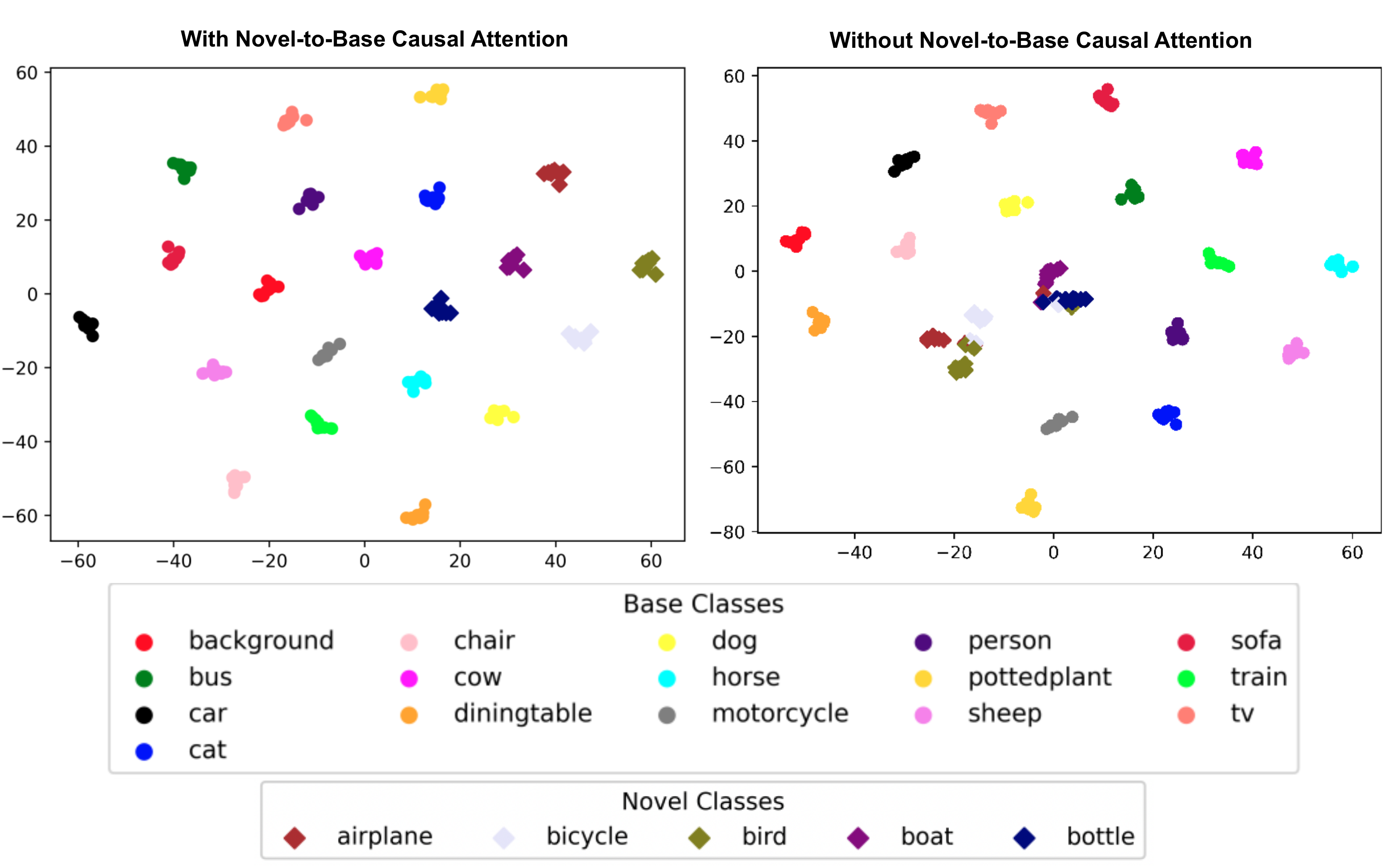}
    \vspace{-0.2in}
    \captionof{figure}{{\bf TSNE visualization of the learned base and novel prompt features:} (left) with Novel-to-Base Causal Attention; (right) without Novel-to-Base Casual Attention. }
    \label{fig:tsne}
    \vspace{-0.10in}
\end{figure}

\subsection{Ablation}
We performed ablation studies (refer to Table~\ref{tab:ablation}) to explain our design choices and contribution of each components. 

\vspace{0.01in}
\noindent \textbf{Visual Prompts Initialization.} As seen in \textbf{(1)} and \textbf{(2)} of the Table~\ref{tab:ablation},  initializing novel prompts with masked mean pooling from the support set images improves both base and novel class accuracy compared to random initialization. This is more pronounced with novel-to-base causal attention~\textbf{(5)}--\textbf{(6)}, amplifying the performance gain from masked pooling initialization. 

\vspace{0.01in}
\noindent \textbf{Novel-to-base Causal Attention.} 
As seen in \textbf{(4)}--\textbf{(7)} of Table~\ref{tab:ablation}, novel-to-base causal attention yields the most significant improvement in novel class accuracy compared to \textbf{(1)}--\textbf{(2)}. This aligns with our hypothesis that uni-directional causal attention between rich base prompts (learned with many examples) and novel prompts (learned with few examples) increases separation among novel classes and pushes the feature space of base prompts away from the novel prompts. It is further confirmed by Figure~\ref{fig:confusion_matrix} and Figure~\ref{fig:tsne}. As observed in Figure~\ref{fig:confusion_matrix}, the novel-to-base causal attention leads to lower confusion between base and novel classes. Figure~\ref{fig:tsne} shows a t-SNE~\cite{van2008visualizing} visualization of the features of base and novel visual prompts for $10$ different test images. As observed, the incorporation of novel-to-base causal attention leads to much clearer separation of visual prompt features, particularly for novel prompts.

\vspace{0.01in}
\noindent \textbf{Multiscale Causal Attention.} We also ablated the placement of the causal attention module. We observe that applying it only at the first layer \textbf{(3)} slightly improves novel mIoU compared to the model without causal attention. Yet, it is notably inferior to applying at every decoder layer \textbf{(4)}-- \textbf{(7)}. We could have opted for separate parameters for the causal attention module at each layer. However, as seen in \textbf{(4)}, having separate weights offers little improvement to novel mIoU despite significant increase in total number of parameters to optimize. In fact, it decreases the base accuracy, leading to overall worse performance. 

\begin{figure}[t]
    \centering
    \includegraphics[width=0.5\textwidth]{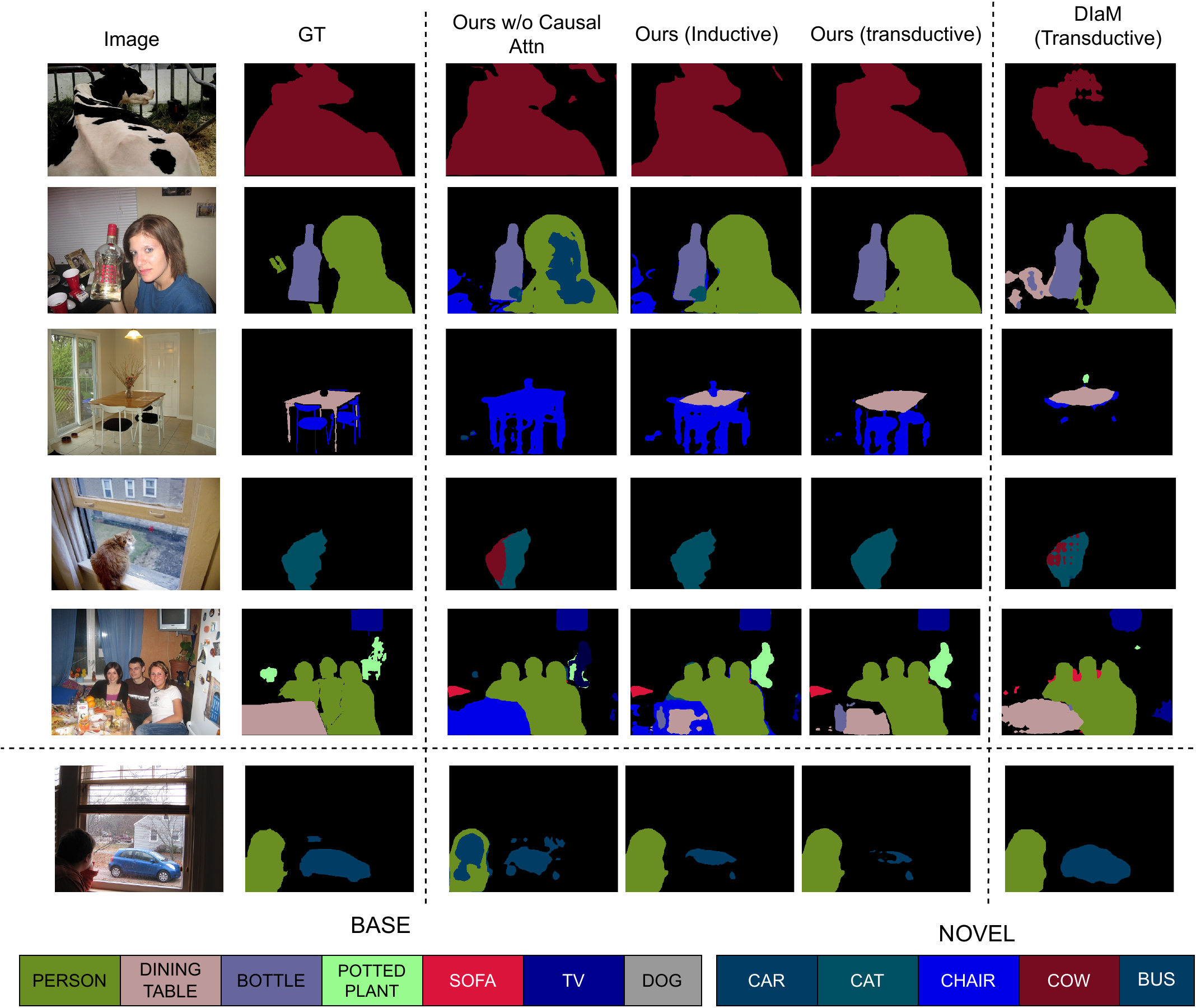}
    \captionof{figure}{{\bf Qualitative Results} for 1-shot on Pascal-$5^i$. The leftomost two columns show image and ground truth mask; (Third) Baseline without causal attention; (Fourth); Ours in inductive setting; (Fifth) Ours in transductive setting; (Last) DIaM~\cite{hajimiri2023strong}. \leon{Last row illustrates a {\em failure}.} }
    \label{fig:qual}
\vspace{-8pt}
\end{figure}
\noindent \textbf{Qualitative Results.} Figure~\ref{fig:qual} compares the qualitative results of our model for 1-shot GFSS on Pascal-$5^i$ in both inductive and transductive settings with the baseline lacking causal attention and DIaM~\cite{hajimiri2023strong}. Notably, for first row, both our inductive and transductive approaches predict more accurate segmentation of the \emph{cow} compared to DIaM. In second row, the causal attention improves \emph{person} segmentation followed by transduction. In the third row, DIaM completely misses the novel category \emph{chair}. The causal attention allows our model to better differentiate the base class \emph{dining table} from the novel class \emph{chairs}. Transduction improves accuracy of the segmentation of \emph{dining table}. In fourth row, we can observe that both DIaM and our model without causal attention misdetects novel class \emph{cat} for \emph{cow} at multiple locations. Causal attention allows our model to make a uniform and more accurate prediction of \emph{cat}. In the fifth row, we observe that DIaM~\cite{hajimiri2023strong} completely misses detecting the base category \emph{potted plant} and poorly segments \emph{person}. The baseline without causal attention misdetects the \emph{potted plant} with novel class \emph{bus} and misdetects \emph{dining table} with \emph{chair}. Causal attention improves the accuracy which is further improved by transduction. Last row demonstrates a case where our model performs worse than DIaM~\cite{hajimiri2023strong}. Again, we observe that adding causal attention reduces misclassification of \emph{person} with \emph{car}, although overall the segmentation of \emph{car} is less accurate than DIaM~\cite{hajimiri2023strong}.


From the qualitative results, the effectiveness of novel-to-base causal attention at distinguishing base from novel classes becomes apparent. Our model in the inductive setting outperforms transductive DIaM in most cases, despite the inference time being faster.

\vspace{-6pt}


\section{Conclusion}

\vspace{-4pt}
In this work, we proposed a technique to learn visual prompts from few-shot examples of novel unseen cateogries to prompt a pre-trained segmentation model that not only generalizes well on the unseen novel class but strongly retains performance over the base classes on which the model was pre-trained. We achieve this by prompting the image features at multiple scales and proposing a uni-directional novel-to-base causal attention mechanism that enriches the novel visual prompts through interaction with the base visual prompts. Our proposed method can be applied in both inductive and transductive settings and achieves state-of-the-art performance on two benchmark few-shot segmentation datasets.
\footnotetext{ \textbf{Acknowledgments and Disclosure of Funding.} This work was partially funded by the Vector Institute for AI, Canada CIFAR AI Chair, NSERC CRC and NSERC DG. Hardware resources were partially provided by the Province of Ontario, the Government of Canada through CIFAR, and \href{https://vectorinstitute.ai/\#partners}{companies} sponsoring the Vector Institute. Additional support was provided by JELF CFI grant and Digital Research Alliance of Canada under the RAC award. }
{
    \small
    \bibliographystyle{ieeenat_fullname}
    \bibliography{main}
}

\title{Supplementary Material of Visual Prompting for Generalized Few-shot Segmentation: A Multi-scale Approach}

\maketitlesupplementary

\setcounter{section}{0}
\setcounter{table}{0}
\setcounter{figure}{0}
\renewcommand{\thesection}{\Alph{section}}
\renewcommand{\thetable}{\Alph{table}} 
\renewcommand{\thefigure}{\Alph{figure}} 


\begin{abstract}
This document provides additional material that is supplemental to our main submission. Section~\ref{sec:efficiency} describes the computational efficiency results. Section~\ref{sec:ablation} includes additional ablation studies. Section~\ref{sec:split_class_results} provides split and class-wise performance results, followed by Section~\ref{sec:multi-scale} that analyzes the performance on objects of varying size. \textcolor{black}{Section~\ref{sec:cross-dataset} discusses the performance of our model in a cross-dataset scenario.} Section~\ref{sec:add-qual} discusses additional qualitative results on COCO-20$^i$ dataset. Finally, Section~\ref{sec:societal} details the societal impact of our work as standard practice in computer vision research.
\end{abstract}

 \section{Computational efficiency comparison} 
 \label{sec:efficiency}
In Table 1 in the main submission, we demonstrated the superior performance of our approach, even in the inductive setting, outperforming transductive methods such as DIaM. Transductive approaches necessitate test-time optimization for each data example computing transductive losses, leading to higher inference times. As illustrated in Table~\ref{tab:params}, there exists approximately 20$\times$ increase in inference speed of our inductive approach compared to transductive methods like DIaM~\cite{hajimiri2023strong}. Another important observation is that while our model has a higher number of parameters overall, it is computationally efficient as the number of FLOPs is approximately 57\% lower. 


 \section{Additional Ablation Studies}
 \label{sec:ablation}
 \subsection{Number of Training Iterations} In Figure~\ref{fig:iters}, we present a graph of how our performance varies in the inductive setting for  different number of training iterations. While the results reported in Table 1 of the main paper were evaluated after 100 training iterations, both inductive and transductive, the figure reveals that extending the optimization process to a higher number of iterations consistently enhances performance, peaking notably at 300 iterations, but at the expense of run-time.

\subsection{Transduction Ablation} As mentioned in Section 4.4 of the original paper, our observations indicate that incorporating transductive losses from the initial iteration results in sub-optimal mIoU on novel classes. This is attributed to inaccurate estimation of the label marginal distribution that is used in the transductive losses. To illustrate this, Figure~\ref{fig:start_trans} showcases the performance of our model in 1-shot inference within a transductive setting, varying the number of iterations at which transduction is applied. Notably, it demonstrates that applying transduction either early or late results in performance degradation.

 \begin{table}[t]
     \centering
     \aboverulesep=0ex
     \belowrulesep=0ex
    \begin{adjustbox}{max width=\columnwidth}
    \begin{tabular}{l|l|c|c|c| c}
    
        \toprule   
    Model & Learning & mean mIoU & Total Params & FLOPs & Inference Time \\
\bottomrule
    Ours & Inductive & \textbf{54.79} & 69.19M & \textbf{55.16G} & \textbf{0.015s} \\
    DIaM~\cite{hajimiri2023strong} & Transductive & 53.00 & \textbf{46.72M} & 128.26G & 0.32s \\
    
        \bottomrule
      \end{tabular}
  \end{adjustbox}
  \captionsetup{labelfont=bf}
     \caption{Parameter and 1-shot inference time comparison on PASCAL-$5^i$ dataset. FLOPs calculation is done for forward pass only and is computed using the flopth library~\href{https://pypi.org/project/flopth/}{https://pypi.org/project/flopth/}. To compute FLOP we used a image size $417 \times 417$. }
     \label{tab:params}
\end{table}

\begin{figure}
  \centering
\resizebox{0.47\textwidth}{!}{
  \begin{tikzpicture}
    \begin{axis}[
      width=0.5\textwidth,
      height=7cm,
      xlabel={Number of Iterations},
      ylabel={mIoU (\%)},
      title={\textbf{mIoU Over Iterations}},
      legend style={at={(0.5,-0.20)}, anchor=north, legend columns=-1},
      xtick={10,100,200,300,400,500},
      ytick={20,30,40,50,60,70,80,90},
      grid=both,
      grid style={dashed,gray!30},
      legend cell align={left},
      xmin=-50
      ]

      \addplot[mark=*, color=blue, solid] coordinates {(10,74.40) (25,74.37) (50,74.78) (75,74.72) (100,74.58) (150,74.68) (200,74.68) (250,74.68) (300,74.70) (400,74.71) (500,74.70)};
      \addlegendentry{Base mIoU}

      \addplot[mark=square*, color=red, solid] coordinates {(10,25.31) (25,31.22) (50,33.54) (75,34.19) (100,34.99) (150,35.43) (200,35.81) (250,36.06) (300,36.23) (400,36.37) (500,36.36)};
      \addlegendentry{Novel mIoU}

      \addplot[mark=triangle*, color=green!30!black, dashed] coordinates {(10,49.85) (25,52.79) (50,54.16) (75,54.46) (100,54.78) (150,55.06) (200,55.24) (250,55.37) (300,55.46) (400,55.54) (500,55.53)};
      \addlegendentry{Mean}
      
\node[above, blue] at (axis cs:10,74.40) {74.4};
\node[above, blue] at (axis cs:100,74.58) {74.6};
\node[above, blue] at (axis cs:200,74.68) {74.7};
\node[above, blue] at (axis cs:300,74.70) {74.7};
\node[above, blue] at (axis cs:400,74.71) {74.7};
\node[above, blue] at (axis cs:500,74.70) {74.7};

\node[above, red] at (axis cs:-5, 25.3) {25.3};
\node[above, red] at (axis cs:45, 33.54) {33.5};
\node[above, red] at (axis cs:100,34.99) {35.0};
\node[above, red] at (axis cs:200,35.18) {35.2};
\node[above, red] at (axis cs:300,36.23) {36.2};
\node[above, red] at (axis cs:400,36.37) {36.4};
\node[above, red] at (axis cs:500,36.36) {36.4};

\node[above, green!30!black] at (axis cs:-5, 50) {49.9};
\node[above, green!30!black] at (axis cs:45, 54.16) {54.2};
\node[above, green!30!black] at (axis cs:100, 54.78) {54.8};
\node[above, green!30!black] at (axis cs:200, 55.24) {55.2};
\node[above, green!30!black] at (axis cs:300, 55.46) {55.5};
\node[above, green!30!black] at (axis cs:400,55.54) {55.5};
\node[above, green!30!black] at (axis cs:500,55.53) {55.5};


    \end{axis}
  \end{tikzpicture}
  }
  \caption{1-shot generalized few-shot segmentation performance for different numbers of training iterations (inductive setting) on Pascal-$5^i$ dataset.}
  \label{fig:iters}
\end{figure}
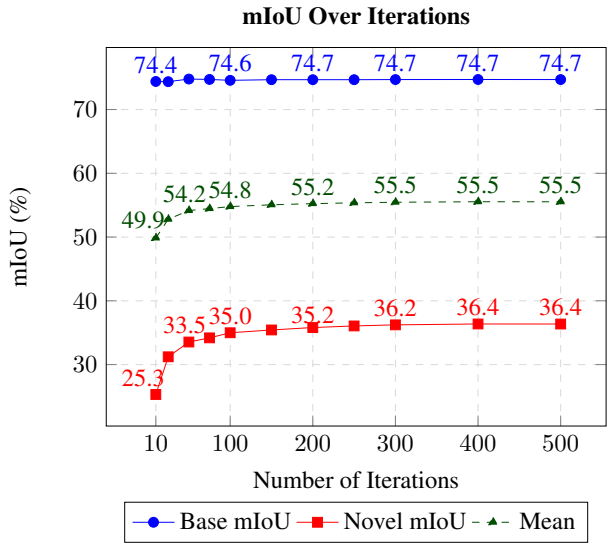

\begin{figure}
  \centering
  \resizebox{0.47\textwidth}{!}{
  \begin{tikzpicture}
    \begin{axis}[
      width=\columnwidth,
      height=6cm,
      xlabel={Number of Iterations},
      ylabel={mIoU (\%)},
      title={\textbf{Impact of Starting Iteration of Transduction}},
      legend style={at={(0.5,-0.25)}, anchor=north, legend columns=-1},
      xtick={0, 20, 40, 70, 90},
      ytick={20, 30, 40, 50, 60, 70, 80, 90},
      grid=both,
      legend cell align={left},
      grid style={dashed,gray!30},
      ymax = 85,
      ymin = 20,
      xmin = -10,
      xmax = 100
      ]

      \addplot[mark=*, color=blue, solid] coordinates {(0,76.36) (20,76.38) (40,76.40) (70,76.38) (90,75.73)};
      \addlegendentry{Base mIoU}

      \addplot[mark=square*, color=red, solid] coordinates {(0,35.14) (20,45.39) (40,47.32) (70,47.12) (90,42.78)};
      \addlegendentry{Novel mIoU}

      \addplot[mark=triangle*, color=green!30!black, dashed] coordinates {(0,55.75) (20,60.88) (40,61.86) (70,61.75) (90,59.26)};
      \addlegendentry{Mean mIoU}

\node[above, blue] at (axis cs:0,76.36) {76.4};
\node[above, blue] at (axis cs:20,76.38) {76.4};
\node[above, blue] at (axis cs:40,76.40) {76.4};
\node[above, blue] at (axis cs:70,76.38) {76.4};
\node[above, blue] at (axis cs:90,75.73) {75.7};

\node[above, red] at (axis cs:0, 35.14) {35.1};
\node[above, red] at (axis cs:20,45.39) {45.4};
\node[above, red] at (axis cs:40,47.32) {47.3};
\node[above, red] at (axis cs:70,47.12) {47.1};
\node[above, red] at (axis cs:90,42.78) {42.8};

\node[above, green!30!black] at (axis cs:0, 55.75) {55.8};
\node[above, green!30!black] at (axis cs:20, 60.88) {60.9};
\node[above, green!30!black] at (axis cs:40, 61.86) {61.9};
\node[above, green!30!black] at (axis cs:70, 61.75) {61.8};
\node[above, green!30!black] at (axis cs:90, 59.26) {59.3};

    \end{axis}
  \end{tikzpicture}
  }
  \caption{1-shot GFSS performance for the number of iterations at which we start applying transduction losses on Pascal-$5^i$ dataset.}
  \label{fig:start_trans} 
\end{figure}
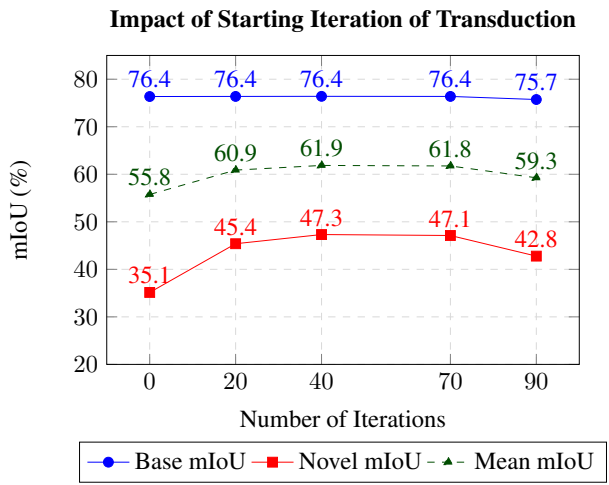

\begin{figure}
    \centering
    \resizebox{0.47\textwidth}{!}{
    \begin{tikzpicture}
        \begin{axis}[
            title={\textbf{Base class shot-wise mIoU}},
            width=\columnwidth,
            height =6cm,
            xlabel={Shots},
            ylabel={mIoU (\%)},
            legend style={at={(0.5,-0.3)},anchor=north,legend columns=-1},
            xtick={1,3,5,7,10},
            ytick={60,65,70,75,80},
            grid=both,
            grid style={dashed,gray!30},
        ]
        
        \addplot[color=blue,mark=*] coordinates {
            (1,74.58)
            (3,74.45)
            (5,74.86)
            (7,75.30)
            (10,74.67)
        };
        \addlegendentry{Ours (inductive)}
        
        \addplot[color=red,mark=square] coordinates {
            (1,69.09)
            (3,68.92)
            (5,68.50)
            (7,68.33)
            (10,68.49)
        };
        \addlegendentry{Ours w/o CA}
        
        \addplot[color=green!30!black,mark=triangle] coordinates {
            (1,66.79)
            (3,64.44)
            (5,64.05)
            (7,63.84)
            (10,63.41)
        };
        \addlegendentry{DIaM (inductive)}

\node[below, blue] at (axis cs:1,74.58) {74.6};
\node[below, blue] at (axis cs:3,74.45) {74.5};
\node[below, blue] at (axis cs:5,74.86) {74.9};
\node[below, blue] at (axis cs:7,75.30) {75.3};
\node[below, blue] at (axis cs:10,74.67) {74.7};

\node[above, red] at (axis cs:1, 69.09) {69.1};
\node[above, red] at (axis cs:3,68.92) {68.9};
\node[above, red] at (axis cs:5,68.50) {68.5};
\node[above, red] at (axis cs:7,68.33) {68.3};
\node[above, red] at (axis cs:10,68.49) {68.5};

\node[above, green!30!black] at (axis cs:1, 66.79) {66.8};
\node[above, green!30!black] at (axis cs:3, 64.44) {64.4};
\node[above, green!30!black] at (axis cs:5, 64.05) {64.1};
\node[above, green!30!black] at (axis cs:7, 63.84) {63.8};
\node[above, green!30!black] at (axis cs:10, 63.41) {63.4};

        \end{axis}
    \end{tikzpicture}
    }
    \caption{Base Class IoU comparison in inductive setting for our approach against both; the baseline without causal attention and inductive DIaM~\cite{hajimiri2023strong}, at various support set shots on PASCAL-$5^i$. w/o CA: without causal attention.}
    \label{fig:shots_base}
\end{figure}
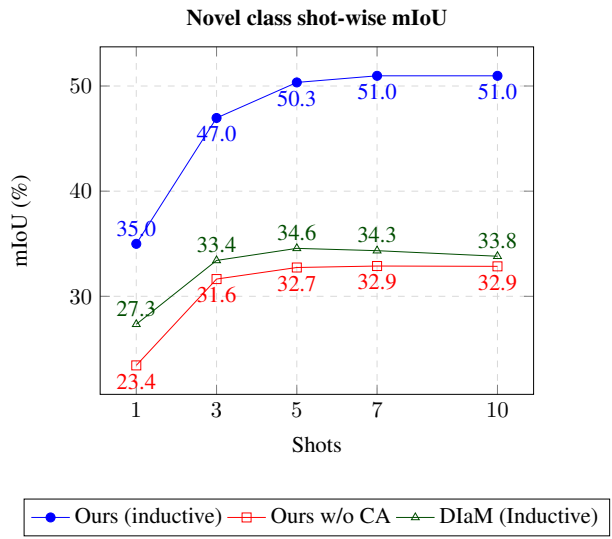
\begin{figure}
    \centering
    \resizebox{0.47\textwidth}{!}{
    \begin{tikzpicture}
        \begin{axis}[
            title={\textbf{Novel class shot-wise mIoU}},
            width=\columnwidth,
            height=7cm,
            xlabel={Shots},
            ylabel={mIoU (\%)},
            legend style={at={(0.5,-0.3)},anchor=north,legend columns=-1},
            xtick={1,3,5,7,10},
            ytick={20,30,40,50,60},
            grid=both,
            grid style={dashed,gray!30},
        ]
        
        \addplot[color=blue,mark=*] coordinates {
            (1,34.99)
            (3,46.96)
            (5,50.34)
            (7,50.97)
            (10,50.97)
        };
        \addlegendentry{Ours (inductive)}
        
        \addplot[color=red,mark=square] coordinates {
            (1,23.43)
            (3,31.64)
            (5,32.74)
            (7,32.88)
            (10,32.85)
        };
        \addlegendentry{Ours w/o CA}
        
        \addplot[color=green!30!black,mark=triangle] coordinates {
            (1,27.34)
            (3,33.41)
            (5,34.56)
            (7,34.34)
            (10,33.81)
        };
        \addlegendentry{DIaM (Inductive)}

\node[above, blue] at (axis cs:1, 34.99) {35.0};
\node[below, blue] at (axis cs:3,46.96) {47.0};
\node[below, blue] at (axis cs:5,50.34) {50.3};
\node[below, blue] at (axis cs:7,50.97) {51.0};
\node[below, blue] at (axis cs:10,50.97) {51.0};

\node[below, red] at (axis cs:1, 23.43) {23.4};
\node[below, red] at (axis cs:3,31.64) {31.6};
\node[below, red] at (axis cs:5,32.74) {32.7};
\node[below, red] at (axis cs:7,32.88) {32.9};
\node[below, red] at (axis cs:10,32.85) {32.9};

\node[above, green!30!black] at (axis cs:1, 27.34) {27.3};
\node[above, green!30!black] at (axis cs:3, 33.41) {33.4};
\node[above, green!30!black] at (axis cs:5, 34.56) {34.6};
\node[above, green!30!black] at (axis cs:7, 34.34) {34.3};
\node[above, green!30!black] at (axis cs:10, 33.81) {33.8};
        
        \end{axis}
    \end{tikzpicture}
    }
    \caption{Base Class performance comparison in inductive setting for our approach against the baseline without causal attention and inductive DIaM~\cite{hajimiri2023strong} at various support set shots on PASCAL-$5^i$. w/o CA: without causal attention.}
    \label{fig:shots_novel}
\end{figure}

\subsection{Shots Analysis} We conducted an ablation to evaluate the performance of our model in an inductive setting, considering various shot configurations. We compared it with the baseline lacking novel-to-base causal attention and inductive DIaM (without transduction)~\cite{hajimiri2023strong}. 

In Figure~\ref{fig:shots_base}, we demonstrate the variation in base class performance across different shots. As observed, our method maintains a consistently high base class performance. In contrast, the model without causal attention exhibits a gradual decline in base class performance as the number of shots increases. \textcolor{black}{Our base class mIoU remains largely flat with increasing shots since it has already converged during large-scale training.} Notably, inductive DIaM experiences a sharp decline in base performance with the growing number of shots. This observation underscores the robustness of our model in retaining base class performance as it encounters more examples of novel classes. 

Likewise, in Figure~\ref{fig:shots_novel}, we illustrate the change of novel class performance across a different number of shots. Notably, our model demonstrates a consistent improvement in novel class performance as the number of shots increases. In contrast, both the baseline without causal attention and inductive DIaM~\cite{hajimiri2023strong} exhibit a plateau in novel class performance around the 5-shot mark. In our model, the leap in performance between different shots is notably more pronounced, demonstrating its capacity for significant improvement with increasing number of examples of the novel class.

\begin{table}[!]
\centering
\vspace{24pt}
\begin{adjustbox}{max width=0.48\textwidth}
\begin{tabu}{@{}llcccccc@{}}
\tabucline[1pt]{-}
&  & \multicolumn{6}{c}{} \\ 
 &  &  \multicolumn{6}{c}{\textbf{Inductive Setting}}\\

\multirow{2}{*}{} & \multirow{2}{*}{} & \multicolumn{3}{c}{1-shot} & \multicolumn{3}{c}{5-shot}\\ 

\cmidrule(r{2pt}){3-5} \cmidrule(r{2pt}){6-8}

Dataset & Split  & Base  & Novel & Mean & Base & Novel & Mean  \\ \hline

\multirow{4}{*}{COCO-20$^i$} & 0  & 50.80  & 13.45  & 32.13 & 50.82 & 25.54  & 38.18   \\ 
  & 1  & 50.04  & 22.07  & 36.06 & 50.62 & 34.35  & 42.49   \\ 

  & 2  & 53.50  & 18.50  & 35.77 & 53.29 & 31.41  & 42.17  \\ 

   & 3  & 51.85  & 17.99  & 34.85 & 51.75 & 28.95  & 40.35   \\ 

      & mean  & 51.55  & 18.00  & 34.78 & 51.59 & 30.06  & 40.80   \\

\bottomrule

\multirow{4}{*}{Pascal-5$^i$} & 0  & 75.95 & 29.86  & 52.91 & 76.70 & 46.72  & 61.71   \\ 
  & 1  & 72.44  & 45.59  & 59.02 & 73.14 & 58.09  & 65.62   \\ 

  & 2  & 72.16 & 34.08  & 53.12 & 72.66  & 56.09 & 64.38 \\ 

   & 3  & 77.75  & 30.43  & 54.09 & 76.95 & 40.45 & 58.70  \\ 

   & mean  & 74.58  & 34.99  & 54.79 & 74.86 & 50.35  & 62.60  \\

\bottomrule

\multirow{1}{*}{} & \multirow{1}{*}{} & \multicolumn{6}{c}{} \\ 
 &  &  \multicolumn{6}{c}{\textbf{Transductive Setting}}\\

 \multirow{2}{*}{} & \multirow{2}{*}{} & \multicolumn{3}{c}{1-shot} & \multicolumn{3}{c}{5-shot}\\ 

\cmidrule(r{2pt}){3-5} \cmidrule(r{2pt}){6-8}

Dataset & Split  & Base  & Novel & Mean & Base & Novel & Mean  \\ \hline

\multirow{4}{*}{COCO-20$^i$} & 0  & 53.59  & 15.25  & 34.42 & 54.68 & 27.55  & 41.12  \\ 
  & 1  & 52.37  & 22.67  & 37.52 & 52.39 & 36.08  & 44.24  \\ 

  & 2  & 54.93  & 18.81  & 36.87 & 55.12 & 31.87  & 43.50  \\ 

   & 3  & 54.31  & 16.48  & 35.40 & 53.05 & 29.05  & 41.05   \\ 

  & mean  & 53.80  & 18.30  & 36.05 & 53.81 & 31.14  & 42.48   \\

\bottomrule

\multirow{4}{*}{Pascal-5$^i$} & 0  & 76.62 & 33.69 & 55.16 & 76.65 & 53.02  & 64.84  \\ 
  & 1  & 75.46  & 50.95  & 63.21 & 75.47 & 63.30  & 69.39  \\ 

  & 2  & 74.64 & 39.65  & 57.15 & 74.67  & 60.51 & 67.59 \\ 

   & 3  & 78.82  & 35.02  & 56.92 & 78.89 & 47.65 & 63.27  \\ 

   & mean  & 76.39 & 39.83  & 58.11 & 76.42 & 56.12  & 66.27   \\

\bottomrule

\end{tabu}
\end{adjustbox}
\captionsetup{labelfont=bf}
\captionof{table}{GFSS performance for each split in inductive \textbf{(top)} and transductive \textbf{(bottom)} settings respectively.}

 \label{tab:splitwise}
\end{table}
\begin{table*}[!]
\centering

 \begin{adjustbox}{max width=\textwidth}
\begin{tabu}{@{}|l|lcc|lcc|lcc|lcc|@{}}
\tabucline[1pt]{-}

\multirow{2}{*}{} & \multicolumn{3}{c|}{Split 0} & \multicolumn{3}{c|}{Split 1} & \multicolumn{3}{c|}{Split 2} & \multicolumn{3}{c|}{Split 3} \\

\multirow{15}{*}{Base Classes} &Class Name & With Causal Attn & No Causal Attn & Class Name & With Causal Attn & No Causal Attn & Class Name & With Causal Attn & No Causal Attn & Class Name & With Causal Attn & No Causal Attn\\ 
\hline

& \textbf{Bus} & 93.85 & 93.80 & \textbf{Airplane} & 88.05 & 82.85 & \textbf{Airplane} & 87.48 & 87.70 & \textbf{Airplane} & 87.81 & 88.26 \\

& \textbf{Car} & 84.27 & 84.10 & \textbf{Bicycle} & 45.10 & 35.21 & \textbf{Bicycle} & 36.84 & 37.93 & \textbf{Bicycle} & 43.33 & 43.34 \\

& \textbf{Cat} & 92.15 & 92.08 & \textbf{Bird} & 85.55 & 77.46 & \textbf{Bird} & 87.51 & 86.39 & \textbf{Bird} & 86.67 & 87.43 \\

& \textbf{Chair} & 36.67 & 30.13 & \textbf{Boat} & 70.93 & 71.28 & \textbf{Boat} & 72.23 & 72.08 & \textbf{Boat} & 75.08 & 75.46 \\

& \textbf{Cow} & 90.67 & 89.80 & \textbf{Bottle} & 81.54 & 59.00 & \textbf{Bottle} & 81.02 & 68.70 & \textbf{Bottle} & 80.06 & 81.31 \\

& \textbf{Dining Table} & 55.20 & 46.43 & \textbf{Dining Table} & 57.99 & 33.90 & \textbf{Bus} & 93.56 & 93.68 & \textbf{Bus} & 92.95 & 89.85 \\

& \textbf{Dog} & 88.74 & 89.03 & \textbf{Dog} & 89.15 & 70.01 & \textbf{Car} & 85.96 & 85.56 & \textbf{Car} & 87.18 & 82.98 \\

& \textbf{Horse} & 84.95 & 83.81 & \textbf{Horse} & 72.94 & 26.94 & \textbf{Cat} & 90.11 & 88.58 & \textbf{Cat} & 92.04 & 92.49 \\

& \textbf{Motorcycle} & 79.45 & 82.50 & \textbf{Motorcycle} & 82.42 & 29.27 & \textbf{Chair} & 30.09 & 30.22 & \textbf{Chair} & 34.20 & 18.14 \\

& \textbf{Person} & 86.56 & 86.61 & \textbf{Person} & 85.81 & 71.86 & \textbf{Cow} & 71.70 & 74.77 & \textbf{Cow} & 88.82 & 88.33 \\

& \textbf{Potted Plant} & 55.47 & 38.44 & \textbf{Potted Plant} & 56.11 & 33.14 & \textbf{Potted Plant} & 53.67 & 49.07 & \textbf{Dining Table} & 57.93 & 58.00 \\

& \textbf{Sheep} & 86.30 & 85.63 & \textbf{Sheep} & 86.19 & 77.59 & \textbf{Sheep} & 84.16 & 64.65 & \textbf{Dog} & 88.57 & 89.62 \\

& \textbf{Sofa} & 43.86 & 42.64 & \textbf{Sofa} & 42.85 & 43.99 & \textbf{Sofa} & 45.54 & 47.11 & \textbf{Horse} & 81.44 & 85.95 \\

& \textbf{Train} & 86.93 & 86.63 & \textbf{Train} & 75.77 & 57.45 & \textbf{Train} & 87.16 & 87.32 & \textbf{Motorcycle} & 84.95 & 82.70 \\

& \textbf{TV} & 75.13 & 71.26 & \textbf{TV} & 73.05 &72.63 & \textbf{TV} & 75.63 & 75.40 & \textbf{Person} & 84.83 & 86.49 \\

\hline

\multirow{5}{*}{Novel Classes} & \textbf{Airplane} & 36.96 & 17.40 & \textbf{Bus} & 51.52 & 35.29 & \textbf{Dining Table} & 13.08 & 13.97 & \textbf{Potted Plant} & 23.69 & 16.37 \\

& \textbf{Bicycle} & 26.95 & 18.02 & \textbf{Car} & 28.98 & 24.65 & \textbf{Dog} & 35.71 & 33.19 & \textbf{Sheep} & 56.30 & 20.32 \\

& \textbf{Bird} & 41.56 & 13.33 & \textbf{Cat} & 74.20 & 43.03 & \textbf{Horse} & 36.77 & 31.11 & \textbf{Sofa} & 12.91 & 13.33 \\

& \textbf{Boat} & 10.15 & 7.25 & \textbf{Chair} & 9.54 & 8.98 & \textbf{Motorcycle} & 48.42 & 49.72 & \textbf{Train} & 38.73 & 33.31 \\

& \textbf{Bottle} & 33.67 & 17.68 & \textbf{Cow} & 63.55 & 38.30 & \textbf{Person} & 36.40 & 31.76 & \textbf{TV} & 20.50 & 16.03 \\

\hline

\end{tabu}
\end{adjustbox}
\captionsetup{labelfont=bf}
\captionof{table}{1-shot GFSS performance for each class in each split of PASCAL-5$^i$ dataset in inductive setting with and without causal attention.}

 \label{tab:classwise}
\end{table*}
\begin{table*}[!]
    \centering
    \begin{adjustbox}{max width=\textwidth}
        \begin{tabu}{@{}lcc|cc|cc|cc|cc@{}}
             \tabucline[1pt]{-}

            \multirow{2}{*}{} & \multicolumn{2}{c|}{Split 0} & \multicolumn{2}{c|}{Split 1} & \multicolumn{2}{c|}{Split 2} & \multicolumn{2}{c|}{Split 3} & \multicolumn{2}{c}{Mean} \\
            \cmidrule(r{2pt}){2-11}
            Object Size & Ours & DIaM~\cite{hajimiri2023strong} & Ours & DIaM~\cite{hajimiri2023strong} & Ours & DIaM~\cite{hajimiri2023strong} & Ours & DIaM~\cite{hajimiri2023strong} & Ours & DIaM~\cite{hajimiri2023strong} \\ 
            \hline

            \textit{Small} & 49.98  & 40.58  & 54.42  & 45.42 & 44.75 & 33.50 & 52.75 & 46.57  & 50.48  & 41.52 \\ 

            \textit{Medium} & 77.47  & 71.29  & 76.63  & 69.00 & 69.66 &54.07 & 76.39 & 72.01  & 75.03  & 66.59 \\ 

            \textit{Large} & 86.78  & 82.53 & 81.68  & 74.91 & 71.68 & 54.98 & 84.78 & 80.34  & 81.23  & 73.19 \\ 
            
            \tabucline[1pt]{-}
        \end{tabu}
    \end{adjustbox}
    \captionsetup{labelfont=bf}
    \caption{Performance analysis of our model in a transductive setting on 1-shot GFSS for \textbf{different object sizes} on PASCAL-5$^i$ compared against DIaM (transductive)~\cite{hajimiri2023strong}. The object sizes are grouped based on the proportion of the image they occupy. Objects occupying more than 30\% of the image are categorized as \emph{large} objects; the objects occupying 10-30\% of the image are classified as \emph{medium}; and rest as \emph{small}.}
    \label{tab:object_size}
\end{table*}

\begin{table}[]
\begin{adjustbox}{max width=0.5\textwidth}
\centering
\begin{tabular}{l|lll}
\toprule
Cross Dataset & Base & Novel & Mean \\ \midrule
Ours (w/o transd.+ causal att.) & 63.8 & 23.9 & 43.9 \\ 
Ours (w/o transd.) & 63.3 & 29.9 & 46.6 \\ 
DiaM~\cite{hajimiri2023strong} & 72.3 & 27.6 & 50.2 \\ 
Ours & \textbf{72.7} & \textbf{32.1} & \textbf{52.2} \\ \bottomrule
\end{tabular}
\end{adjustbox}
\caption{Cross dataset evaluating mIoU (COCO2PASCAL).}
\label{table:coco2pascal}
\end{table}

\begin{figure}
  \centering

  \begin{tikzpicture}
    \begin{axis}[
      width=0.4\textwidth,
      xmin=10,
      xmax=30,
      xlabel={$\sigma$},
      ylabel={mIoU (\%)},
      title={\textbf{Prompt Quality}},
      legend style={at={(0.5,-0.35)}, anchor=north, legend columns=1},
      grid=both,
      grid style={dashed,gray!30},
      legend cell align={left},
      ]

      \addplot[mark=*, color=blue, solid] coordinates {(10, 32.66) (15, 31.97) (20, 31.41) (25, 31.2) (30, 30)};
      \addlegendentry{Additive Noise}
    
      \addplot[color=black, dashed] coordinates {(0, 33.69) (30, 33.69)};
      \addlegendentry{No Additive Noise}
        
    \end{axis}
  \end{tikzpicture}

  \caption{1-shot GFSS performance on split 0 for different additive noise variations (standard deviation $\sigma$ of Gaussian noise) on Pascal-$5^i$ dataset. The additive noise is applied on the support set images which affect the quality of the novel prompts after performing masked average pooling.}
  \vspace{-0.05in}
  \label{fig:rebuttal_noise}
\end{figure}
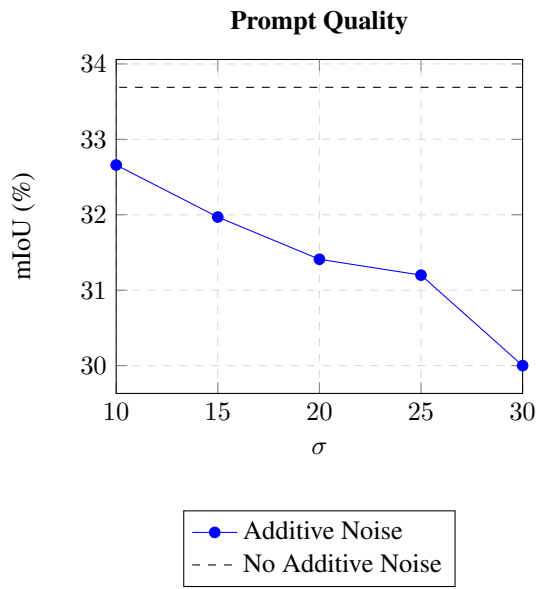

\subsection{Quality of Input Prompts} We conducted an ablation on how the quality of the input prompts affect the overall GFSS performance. For this, we corrupted the input support images, \textcolor{black}{that is used to initialize the novel prompts,} with different levels of additive Gaussian Noise. Figure~\ref{fig:rebuttal_noise} shows the 1-shot GFSS performance for split 0 of Pascal-$5^i$ dataset for different degree of additive Gaussian Noise. As can be observed, there is minimal degradation of performance, even in cases of higher distortions of additive Gaussian Noise with standard deviation of $\sigma = 30$.

\section{Split and Class-wise Results}
\label{sec:split_class_results}
\paragraph{Split-wise Results.} Adhering to the established few-shot segmentation protocol, our model undergoes evaluation across four distinct splits or folds for both the PASCAL-5$^i$ and COCO-20$^i$ datasets. Within each fold or split, a subset of the classes are reserved as novel, serving as the validation set. Results presented in the main manuscript represent an average across all four splits. For a detailed breakdown, Table~\ref{tab:splitwise} provides the performance of our model on each of the four splits individually for both datasets, in both inductive and transductive settings. As expected, the performance on novel classes improves substantially for 5-shot cases across all the splits. Additionally, transduction generally enhances both base and novel class accuracy, except for split-3 of COCO-20$^i$, where the novel class performance experiences a minor degradation, although the base class performance increases.

\begin{figure}[t]
    \centering
    \includegraphics[width=0.5\textwidth]{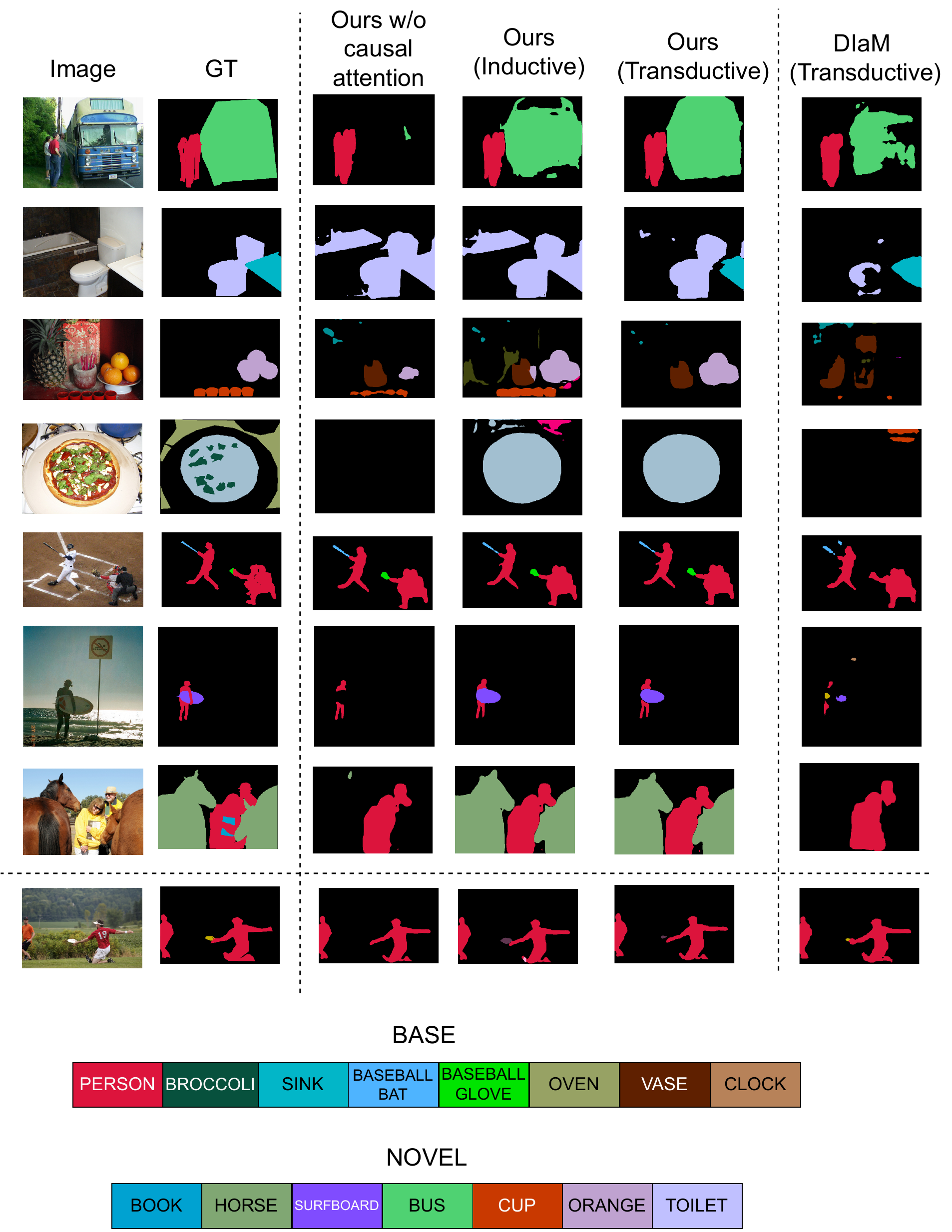}
    \captionof{figure}{{\bf Qualitative Results} for 1-shot GFSS on COCO-20$^i$. The leftomost two columns show image and ground truth mask; (Third) Baseline without causal attention; (Fourth); Ours in inductive setting; (Fifth) Ours in transductive setting; (Last) DIaM~\cite{hajimiri2023strong}. \leon{Last row illustrates a {\em failure}.} }
    \label{fig:qual_supp}
\vspace{-8pt}
\end{figure}

\paragraph{Class-wise Results.} 

In Table~\ref{tab:classwise}, we present the base mIoU and novel mIoU, calculated as the mean across all base and novel classes, respectively. The overall mean mIoU of the Generalized Few-Shot Segmentation (GFSS) is determined by averaging these values, in accordance with the evaluation protocol outlined in~\cite{hajimiri2023strong}. Table~\ref{tab:classwise} provides insights into the 1-shot performance of our model within an inductive setting for each class (both base and novel) on the PASCAL-5$^i$ dataset, covering all four splits.

To emphasize the significance of our model's novel-to-base causal attention, we conduct a comparative analysis against the baseline without causal attention. The results, as depicted in the table, underscore a notable enhancement in performance for both novel and base classes across the various splits. Notably, on split-1, our base class performance exhibits a substantial improvement across most base classes.

\section{Multi-scale Analysis} 
\label{sec:multi-scale}
One of the key components of our approach involves prompting at multiple scales of the image. Our hypothesis posits that this strategy contributes to improved segmentation accuracy across objects of varying sizes. To substantiate this claim, Table~\ref{tab:object_size} presents an analysis of our model's performance in a transductive setting, evaluating objects of different sizes across various splits of Pascal-$5^i$, and comparing it with the transductive DIaM~\cite{hajimiri2023strong}. We categorize objects into three size categories: \emph{small}, \emph{medium}, and \emph{large}, based on the proportion of the image each object occupies. Specifically, we classify objects that occupy more than 30\% of the image as \emph{large}, those occupying between 10-30\% as \emph{medium}, and the remainder as \emph{small}. As depicted in the table, our approach consistently outperforms transductive DIaM~\cite{hajimiri2023strong} across objects of all sizes, notably excelling on small-sized objects.

\section{Cross-Dataset Evaluation}
\label{sec:cross-dataset}  To evaluate the performance of our proposed approach across different domains, we conduct an experiment where the base training is done on split-0 of COCO-$20^i$ and the few-shot inference is performed on Pascal-$5^i$ on classes not overlapping with the base classes of split-0. This non-overlapping categories are following six classes: airplane, boat, chair, dining table, dog and person. Table~\ref{table:coco2pascal} shows the performance of our approach on this cross-domain experiment compared to transductive DIaM~\cite{hajimiri2023strong}, our model in inductive setting, and our model without novel-to-base causal attention. As observed in the table, we substantially outperform DIaM~\cite{hajimiri2023strong} in this cross-dataset experiment, particularly obtaining superior performance on novel categories. Even our inductive setting obtains better novel mIoU than DIaM~\cite{hajimiri2023strong}. Additionally, using our proposed novel-to-base causal attention mechanism leads to a better disentanglement of novel prompts to base prompts, leading to significant improvement in novel mIoU.  

\section{Additional Qualitative Results}
\label{sec:add-qual}
In Figure 5 of the main manuscript, we presented the qualitative results for 1-shot Generalized Few-Shot Segmentation (GFSS) using our model on PASCAL-5$^i$ in both inductive and transductive settings. We compared our outcomes against the baseline lacking causal attention and transductive DIAM~\cite{hajimiri2023strong}. However, no results were shown for COCO-20$^i$. Therefore, in Figure~\ref{fig:qual_supp}, a similar qualitative analysis is presented for 1-shot GFSS on COCO-20$^i$.

As observed, in the first image, the baseline without novel-to-base causal attention fails to appropriately segment the novel class \emph{bus}. The model with causal attention can segment the \emph{bus} with higher accuracy, further improved by transduction. In comparison, the segmentation quality of \emph{bus} in transductive DIAM~\cite{hajimiri2023strong} is considerably worse. In the second image, the baseline without causal attention can detect the novel category \emph{toilet} but misclassifies \emph{sink} and other unlabeled pixels (the tub) as a \emph{toilet}. Adding novel-to-base attention slightly improves segmentation quality but is still unable to correctly classify the \emph{sink}. However, adding transduction loss helps our model correctly classify it. Although DIAM can correctly classify \emph{sink}, it mostly misses the novel category \emph{toilet}. Similarly, in the third image, DIAM completely misses novel classes \emph{orange} and \emph{cup}. Our baseline without causal attention can identify both but with poor accuracy. Our model in the inductive setting can segment both the novel classes \emph{orange} and \emph{cup} with a higher degree of accuracy. Our model in the transductive setting, however, performs worse as it misses \emph{cup}. In the fourth image, both DIAM and the baseline without causal attention miss the novel class \emph{pizza}, and base classes \emph{broccoli} and \emph{oven}. Our models in both inductive and transductive settings can segment \emph{pizza} but fail to identify \emph{broccoli} and \emph{oven}. The fifth image showcases the importance and strength of multi-scale prompting. All our models can detect the small object \emph{baseball gloves} (base class) which DIAM misses. Moreover, DIAM performs worse in segmenting the base class \emph{baseball bat}. Similarly, DIAM and the baseline without causal attention miss novel classes \emph{surfingboard} and \emph{horse} completely in the sixth and seventh images, respectively. Our models (in inductive and transductive settings) can correctly segment them with good accuracy. However, in the seventh image, all methods fail to identify the novel class \emph{book}. Finally, we show a case where DIAM outperforms us in correctly segmenting the novel category \emph{frisbee}. Our models (inductive and transductive) incorrectly classify \emph{frisbee} as \emph{kite}.

Overall, these qualitative results demonstrate the strong performance of our model on both base and novel classes in both inductive and transductive settings. They also underscore the importance of the novel-to-base causal attention module and the multi-scale prompting approach.

\section{Societal Impact}
\label{sec:societal}
Few-shot object segmentation has multiple positive societal impacts as it can be used for a variety of useful applications, \eg, robot manipulation, augmented/virtual reality and assistive technologies (\eg, aid to the blind and low-level vision)~\cite{massiceti2021orbit}. 
It can also help in democratizing computer vision research by  enabling low resourced communities to use the technology, \eg, Africa to develop their own techniques with the limited labelled data available.

However, as with many AI abilities, few-shot object segmentation also can have negative societal impacts. 
Nonetheless, we strongly believe these misuses are available in both few-shot and non few-shot methods and are not tied to the specific few-shot case. On the contrary, we argue that empowering developing countries towards decolonizing artificial intelligence is critical towards a decentralized and ethical AI approach.




\end{document}